\documentclass[11pt,a4paper]{article}
\usepackage{authblk}

\usepackage[hyperref]{acl2020}
\usepackage{times}
\usepackage{latexsym}

\usepackage{array,multirow,graphicx}
\usepackage{float}
\usepackage{amsmath}
\usepackage{caption}
\usepackage{subcaption}
\usepackage{amssymb}
\usepackage{multirow}

\usepackage{supertabular,booktabs}

\usepackage{microtype}

\aclfinalcopy % Uncomment this line for the final submission
\title{Autoencoding Word Representations through Time for Semantic Change Detection}

\author[1,2]{\textbf{Adam Tsakalidis}}
\author[1,2,3]{\textbf{Maria Liakata}}
\vspace{-1cm}
\affil[1]{Queen Mary University of London, London, UK}
\affil[2]{The Alan Turing Institute, London, UK}
\affil[3]{University of Warwick, Coventry, UK}
\affil[ ]{\texttt{\{atsakalidis, mliakata\}@turing.ac.uk}}

\date{}

\begin{document}
\maketitle
\begin{abstract}
    Semantic change detection concerns the task of identifying words whose meaning has changed over time. 
    %With words represented via a distributional semantic model over different periods in time, the current state-of-the-art ignores the temporal aspect of the task by aligning these representations over different time periods in a pairwise fashion and measuring the semantic change level of a word by means of its displacement error that results from the alignmenttreating it instead as a similarity task of the word representations across distinct chunks. 
    The current state-of-the-art detects the level of semantic change in a word by comparing its vector representation in two distinct time periods, without considering its evolution through time.
    %the relation    between consecutive time spans. 
    In this work, we propose three variants of %seq2seq 
    sequential models for detecting semantically shifted words, effectively accounting for the changes in the word representations over time, in a temporally sensitive manner. Through extensive experimentation under various settings with both synthetic and real data we showcase the importance of sequential modelling of word vectors through time for detecting the words whose semantics have changed the most. Finally, we take a step towards comparing different approaches in a quantitative manner, demonstrating that the temporal modelling of word representations yields a clear-cut advantage in performance.
\end{abstract}

\section{Introduction}
Identifying words whose lexical meaning has changed over time is a primary area of research at the intersection of natural language processing and historical linguistics. Through the evolution of language, the task of ``semantic change detection'' \cite{tang2018state} can provide valuable insights on cultural evolution over time \cite{michel2011quantitative}. Measuring linguistic change more broadly is also relevant to understanding the dynamics in online communities \cite{Leskovec13a} and the evolution of individuals, e.g. in terms of their expertise \cite{Leskovec13b}. Recent years have seen a surge in interest in this area since researchers are now able to leverage the increasing availability of historical corpora in digital form and develop algorithms that can detect the shift in a word's meaning through time.

However, two key challenges in the field still remain. (a) Firstly, 
%there is a gap in existing literature on model comparison. 
there is little work in existing literature on model comparison \cite{schlechtweg2019wind,dubossarsky2019time,shoemark2019broom}. Partially due to the lack of labelled datasets, existing work assesses model performance primarily in a qualitative manner, without comparing results against prior work in a quantitative fashion. Therefore, it becomes impossible to assess \textit{what constitutes an appropriate approach for semantic change detection}. (b) Secondly, on a methodological front, a large body of related work detects semantically shifted words by pairwise comparisons of their representations in distinct periods in time, \textit{ignoring the sequential modelling aspect of the task} \cite{hamilton2016diachronic,tsakalidis2019mining}. Since semantic change is a time-sensitive process \cite{tsakalidis2019mining}, considering intermediate vector representations in consecutive time periods can be crucial to improving model performance~\cite{shoemark2019broom}. This type of modelling approach is very different from considering changes between two distinct bins of word representations  \cite{schlechtweg2018diachronic,schlechtweg2020semeval}.

Here we tackle both of the above challenges by approaching semantic change detection as an anomaly identification task. We propose an encoder-decoder architecture for learning word representations across time. We hypothesize that once such a model has been successfully trained on temporally sensitive word sequences it will be able to accurately predict the evolution of the semantic representation of any word through time. Words that have undergone semantic change will be exactly those that yield the highest errors by the prediction model. Specifically we make the following contributions: 

\begin{itemize}
    \item we develop three variants of an LSTM-based %seq2seq
    neural architecture which enable us to measure the level of semantic change of a word by tracking its evolution through time in a sequential manner. These are: (a) a current word representation autoencoder, (b) a future word representation decoder and (c) a hybrid approach combining (a) and (b);
    \item we showcase the effectiveness of the proposed models under thorough experimentation with synthetic data;
    \item we compare our models against current practices and competitive baselines using real-world data, demonstrating important gains in performance and highlighting the importance of sequential modelling of word vectors across time.
    %\item we release our code and models, to help set up a benchmark for model comparison within the domain in a quantitative fashion. 
\end{itemize}

%Treating the task as a word ranking problem, through extensive experimentation with synthetic and real data, we showcase the effectiveness of our proposed models and demonstrate that they yield better results than existing work by a clear margin. 
%Finally, by releasing our resources for our models and the baselines, we aim at setting up a basis for model comparison in future work in the domain in a quantitative fashion. 

%This paper is organised as follows: first we summarise the related work (section~\ref{sec:literature}) and then we present our models (section~\ref{sec:methodology}). The next two chapters aim at showcasing the appropriateness of the proposed models through experiments with synthetic data (section~\ref{sec:synthetic}) and their comparison against state-of-the-art and other baselines under real-data experimentation (section~\ref{sec:experiments}). Finally, section~\ref{sec:conclusion} concludes with pointers to future directions.

\section{Related Work} \label{sec:literature}
One can distinguish two directions within the literature on semantic change \cite{tang2018state,kutuzov2018diachronic}: (a) learning word representations over discrete time intervals and comparing the resulting vectors and (b) jointly learning the  (diachronic) word representations across time \cite{bamler2017dynamic,rosenfeld2018deep,yao2018dynamic,rudolph2018dynamic}. In this work, we focus on (a) due to scalability issues in (b) associated with learning diachronic representations from very large corpora. Our methods, presented in Section~\ref{sec:methodology}, are applicable to any type of pre-trained word vectors across time.

Related work in (a) derives word representations $W_i$, $i\in [0,..,|T-1|]$ across $|T|$ different time intervals and performs pairwise comparisons for different values of $i$. Early work used frequency- and co-occurrence-based representations for $W_i$ \cite{sagi2009semantic,cook2010automatically,gulordava2011distributional,mihalcea2012word}; however, word2vec-based representations \cite{mikolov2013distributed} has been the standard practice in recent years. Due to the stochastic nature of word2vec, Orthogonal Procrustes (OP) is often firstly applied to the resulting vectors, aiming at aligning the pairwise representations \cite{kulkarni2015statistically,hamilton2016diachronic,del2019short,shoemark2019broom,tsakalidis2019mining,schlechtweg2019wind}. Given two word matrices $W_k$, $W_j$ at times $k$ and $j$ respectively, OP finds the optimal transformation matrix  $R=\underset{\Omega; \Omega^T\Omega=I}{arg min}\left \| \Omega W_k-W_j \right \|_F$ and the semantic shift level of a word $w$ during the time interval $[k,j]$ is defined as the cosine distance $cos(Rw_j, w_k)$ \cite{hamilton2016diachronic}. To tackle the drawback of basing the alignment of the matrices on the whole vocabulary, which assumes that the vast majority of the words remain stable across time, \citet{tsakalidis2019mining} learn the alignment based only on a few semantically stable words across time. However, both approaches operate in a linear  pairwise fashion, thus ignoring the time-sensitive, sequential and possibly non-linear nature of semantic change.

By contrast, \citet{kim2014temporal}, \citet{kulkarni2015statistically} and \citet{shoemark2019broom} derive time series of a word's level of semantic change and use those to detect semantically shifted words. Even though these methods incorporate temporal modelling, they still rely heavily either on the linear transformation $R$ \cite{kulkarni2015statistically,shoemark2019broom} or on the similarity of a word with itself across time via continuous representation learning \cite{kim2014temporal}. The latter has recently been demonstrated to lead to worse performance \cite{shoemark2019broom}. %To this end, a key contribution of our work is that we do not base our methodology on pre-defined transformations to estimate the level of semantic change of a word, but instead propose a non-linear model for learning the word representations across time, effectively exploiting the full sequence of the word's evolution. 

Finally, the comparative evaluation of semantic change detection models is still in its infancy. Most related work assesses model performance based either on an artificial task \cite{rosenfeld2018deep,shoemark2019broom} or on a few hand-picked examples \cite{del2019short}, without cross-model comparison.  Setting a benchmark for model comparison with real-world and sequential word representations would be of great importance to the field.

\section{Methods}\label{sec:methodology}

%\paragraph{Notation} We denote the pre-trained dense representations of the words in a vocabulary $V$ at a particular point in time $t$ with $W_t$. Subsequently, $w^{(a)}_t$ indicates the dense representation of word $a$ at time $t$. Note that the subscript $t$ serves as the index of a particular period in time (e.g., year 2000) and \textit{not} the position of a word in a document.

%We base our methodology on an anomaly detection front. 
We formulate semantic change detection as an anomaly detection task. We hypothesize that the pre-trained word vectors $W_t$ $\in$ $ [W_0$, ..., $W_{|T-1|}]$, where $W_{t}\in\mathbb{R}^{|V|\times d}$ ($|V|$: vocabulary size; $d$: word representation size) in a historical corpus over $|T|$ time periods, evolve according to a non-linear function $f(W_t)$.\footnote{Note that $t$ in $W_t$ represents the time period from when the associated word vectors are taken (e.g., the year 2000) and not the position of a word in a sentence.} By providing an approximation for $f$, we obtain the level of semantic shift of a word $w$ at time $t$ by measuring the distance between its word representation $w_{t}$ against $f(w_{t})$. A key novelty of our work is that we approximate $f$ via a temporally sensitive model using a deep neural network architecture. \citet{shoemark2019broom} showed that accounting for the full sequence of word vectors $[W_0, ..., W_{|T-1|}]$ is more appropriate for detecting semantically shifted words, compared to accounting only for the first and the last representations $[W_0, W_{|T-1|}]$, as is the practice in most earlier work. Following \citet{shoemark2019broom}  we model word evolution by accounting for all intermediate representations across time. 

Our modelling of the semantic change function $f$ is based on two components: (a) an autoencoder, which aims to reconstruct a word's trajectory up to a given point in time $i$  $[w_0, ..., w_{|i|}]$ (section~\ref{sec:methods_past}); and (b) a future predictor, which aims to predict future representations of the word  $[w_{|i+1|}, ..., w_{|T-1|}]$ (section~\ref{sec:methods_future}). The two models can be trained either individually or (c) in combination, in a multi-task setting (section~\ref{sec:methods_multi}).

\begin{figure}[h]
\centering
\includegraphics[width=.85\linewidth]{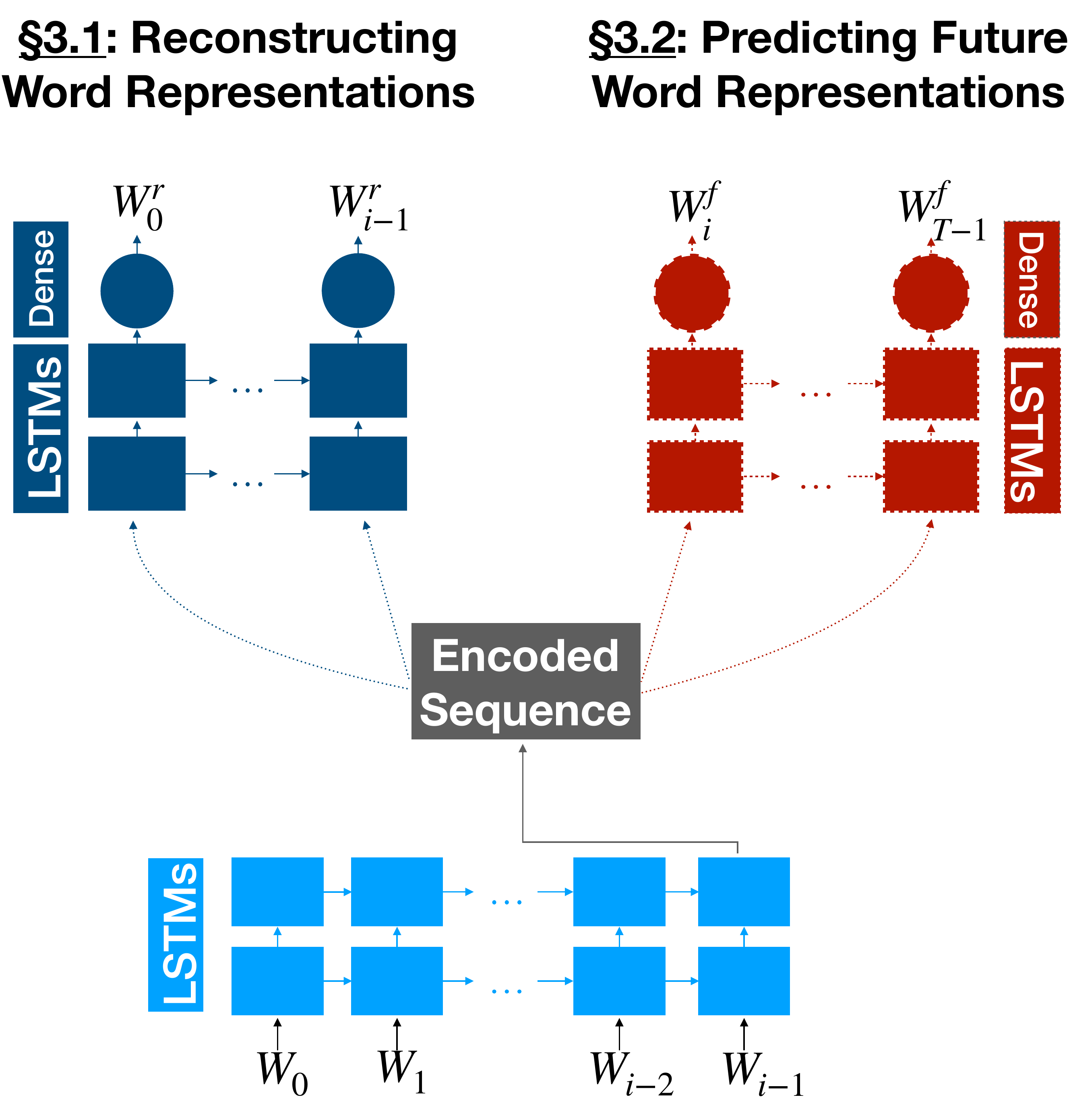}
\caption{Overview of our proposed model: the sequence of the representation of a set of word vectors (Vocabulary) over different time steps $W_{0:i-1}$ is encoded through two LSTM layers and then passed over to a reconstruction (\ref{sec:methods_past}) and a future prediction decoder (\ref{sec:methods_future}). The model is trained by utilising \textbf{either decoder} in isolation, or \textbf{both} of them in parallel (\ref{sec:methods_multi}).}
\label{fig:model}
\end{figure}

\subsection{Reconstructing Word Representations} \label{sec:methods_past}
Given an input sequence of vectors representing the Vocabulary across $i$ points in time $W_{0:i-1} = [W_0, W_1, ..., W_{i-1}]$, the goal of the autoencoder is to reconstruct the input sequence $W_{0:i-1}$, by minimising some loss function. Since the task of semantic change includes a natural temporal dimension, we model our autoencoder via a RNN architecture (see Figure~\ref{fig:model}). The encoder is composed of two LSTM layers \cite{hochreiter1997long} with Dropout layers operating on their outputs, for regularisation \cite{srivastava2014dropout}. The first layer encodes the input sequence of $W_{0:i-1}$ and returns the hidden states to be fed as input to the second layer. The output of the second layer is the final encoded state, which is then copied $|i|$ times and fed as input to the decoder. The decoder has the exact same architecture as the encoder, albeit with additional dense layers on top of the second LSTM layer, fed with the hidden states of the latter, to make the final reconstruction $W^r_{0:i-1}$ on the $|i|$ time steps. The model is trained by minimising the mean squared error (MSE) loss function:\vspace{-.3cm}
\begin{equation}\label{eq:past}
    L_r = \frac{1}{i}\sum_{j=0}^{i-1} (W_j-W^r_j)^2.
\end{equation}
After training, the words that yield the highest error rates in a given test set of word representations through time are considered to be the ones whose semantics have changed the most during the given time period. This assumption is in line with prior work based on word alignment \cite{hamilton2016diachronic,tsakalidis2019mining}, where the alignment error of a word indicates its level of semantic change.

\subsection{Predicting Future Word Representations} \label{sec:methods_future}
Reconstructing the input sequence of word vectors can reveal which words have changed their semantics in the past (i.e., up to time $i-1$, see section~\ref{sec:methods_past}). If we are interested in predicting changes in the semantics of future word representations (i.e., word vectors after time $i-1$), then we can set up a future word representation prediction task, based on a sequence-to-sequence architecture. Formally, given the sequence of past word representations $W_{0:i-1} = [W_0, W_1, ..., W_{i-1}]$ over the first $i$ time points, we want to predict the future representations of the words in the vocabulary $W_{i:T-1} = [W_i, W_{i+1}, ..., W_{T-1}]$, for a sequence of overall length $|T|$ (see Figure~\ref{fig:model}). We follow the same model architecture as described in section~\ref{sec:methods_past}, with the only difference being the number of time steps ($T-i$) that are used in the decoder in order to make $|T-i|$ predictions. The model is trained using the MSE loss function $L_f$:\vspace{-.2cm}
\begin{equation}\label{eq:future}
    L_f = \frac{1}{T-i}\sum_{j=i}^{T-1} (W_j-W^f_j)^2.
\end{equation}

\subsection{Joint Model} \label{sec:methods_multi}
The two models can be combined into a joint one, where, given an input sequence of representations of the vocabulary $W_{0:i-1}$ over $i$ points in time, the goal is both to (a) reconstruct the input sequence and (b) predict the future word $|T-i|$ representations $W_{i:T-1}$. The complete model architecture is provided in Figure~\ref{fig:model}: the encoder is identical to the one used in \ref{sec:methods_past} and \ref{sec:methods_future}. However, the bottleneck is now copied $|T|$ times and passed to the decoders of the reconstruction ($|i|$ times) and future prediction ($|T-i|$ times) components. The loss function $L_{rf}$ used to tune the model parameters is the summation of Eq.~\ref{eq:past} and ~\ref{eq:future}:\vspace{-.2cm}
\begin{equation}\label{eq:multi}
    L_{rf} = \frac{1}{i}\sum_{j=0}^{i-1} (W_j-W^r_j)^2 + \frac{1}{T-i}\sum_{j=i}^{T-1} (W_j-W^f_j)^2.
\end{equation}
There are two main reasons for modelling semantic change in this multi-task setting. Firstly, we benefit from the finer granularity of the two decoders due to their handling of only part of the sequence in a more fine-grained manner, compared to the individual task models. Secondly, the joint model is insensitive to the value of $i$ in Eq.~\ref{eq:multi} compared to Eq.~\ref{eq:past} and \ref{eq:future}. We provide more details on this aspect in \ref{sec:equivanelt}.

\subsection{Model Equivalence}
\label{sec:equivanelt}
The three models perform different operations; however, setting the operational time periods appropriately in  Eq.~\ref{eq:past}-\ref{eq:multi} can result in model equivalence. Specifically, to detect the words whose semantics have changed during [0, $T-1$], the autoencoder in Eq.~\ref{eq:past} needs to be fed and reconstruct the full sequence across [0, $T-1$] (i.e., $i$=$T$-1). Reducing this interval (reducing $i$) would limit the autoencoder's operational time period. On the other hand, an increase in the value of $i$ in Eq.~\ref{eq:future} of the future prediction component shortens the time period during which it can detect the words whose semantics have changed the most -- to account for the whole sequence (i.e., [1, $T-1$]), the future prediction model requires only the $W_0$ word representations in the first time interval to then detect the words whose semantics have changed within [1, $T-1$]. Therefore, setting the parameter $i$ can be crucial for the performance of the two individual models. By contrast, the joint model in section~\ref{sec:methods_multi} is able to detect the words that have undergone semantic change, regardless of the value of $i$ (see Eq.~\ref{eq:multi}), since it is still able to operate on the full sequence -- we showcase these effects in section~\ref{sec:results}.

\section{Experiments with Synthetic Data}\label{sec:synthetic}
In this section we explore the three proposed models and their ability to detect words that have undergone semantic change on an artificial dataset. Tasks ran on artificial data have been used in recent work for evaluation purposes \cite{shoemark2019broom}. We work with artificial data in the current section as a proof-of-concept of our proposed models -- we compare against state-of-the-art models and other baseline methods with real-world data in the following sections. In particular, here we employ a longitudinal dataset of word representations  (\ref{sec:dataset1}) and artificially alter the representations of a small set of words across time (\ref{sec:artificialexamples}). We then train  (\ref{sec:artificialtask}) our models and evaluate them on the basis of their ability to identify those words that have undergone (artificial) semantic change (\ref{sec:synthetics_results}).

\subsection{Dataset}\label{sec:dataset1}
We make use of the UK Web Archive dataset introduced by \citet{tsakalidis2019mining}, which contains 100-dimensional  representations of 47.8K words for each year in the period 2000-2013. These were generated by employing word2vec (i.e., skip-gram with negative sampling)\cite{mikolov2013distributed} on the documents published in each year independently. Each year corresponds to a time step in our modelling. The dataset contains 65 words whose meaning is known to have changed during the same time period as indicated by the Oxford English Dictionary. These are removed for the purposes of this section, to avoid interference with the artificial data modeling. 
We use one subset (80\%) of the remaining word representations across time for training our models and the rest (20\%) for evaluation purposes.

\subsection{Artificial Examples of Semantic Change}\label{sec:artificialexamples}
We generate artificial examples of words with changing semantics, by following a paradigm inspired by \citet{rosenfeld2018deep}. We uniformly at random select 5\% of the words in the test set to alter their semantics. For every selected ``source'' word $\alpha$, we select a ``target'' word $\beta$. Details about the selection process of the target words are provided in the next paragraph. We then alter the representation $w_{t}^{(\alpha)}$ of the source word at each point in time $t$ so that it shifts towards the representation $w_{t}^{(\beta)}$ of the target word at this point in time as: \vspace{-.2cm}
\begin{equation}\label{eq:lambda}
w_{t}^{*(\alpha)} = \lambda_t w_{t}^{(\alpha)}  + (1-\lambda_t) w_{t}^{(\beta)}.
\end{equation}
In our modelling, $\lambda_t$ receives values between 0 and 1 and acts as a decay function that controls the speed of the change in the source word's semantics towards the target. As in \citet{rosenfeld2018deep}, we model $\lambda_t$ via a sigmoid function. Thus, the semantic representation of the word $\alpha$ is not altered during the first time points and then it gradually shifts towards the representation of word $\beta$ (for middle values of $t$), where it stabilizes towards the last time points. Since the duration of the semantic shift of a word may vary, we experiment under three different scenarios, as presented below. 

Different modelling approaches of (artificial) semantic change have been presented in \citet{shoemark2019broom} -- e.g., forcing a word to acquire a new sense while also retaining its original meaning. Here we opted for the ``stronger'' case of semantic shift in Eq.~\ref{eq:lambda} as a proof of concept for our models. In the next section we experiment with uncontrolled (real-world) examples of semantic change, without the need for any hypothesis on the underlying function.

\vspace{.2cm}
\noindent\textbf{Conditioning on Target Words} \hspace{.15cm}
\label{sec:conditioning1}
The selection of the target words should be such that they allow the representation of the source word to change through time. This will not be the case if we select a pair of $\{\alpha,\beta\}$ \{source, target\} words whose representations are very similar (e.g., synonyms). Thus, for each source word $\alpha$ we select uniformly at random a target word $\beta$ s.t. the cosine similarity of their representations at the initial time point $t=0$ (i.e., in year 2000) falls within a certain range $(c-0.1,c]$. Higher values of $c$ enforce a lower semantic change level for $\alpha$ through time, since its representation will be shifted towards a similar word $\beta$, and vice versa. To assess the performance of our models across different semantic change levels, we experiment with varying values for $c$: \{0.0, 0.1, ..., 0.5\}.

\begin{figure}[h]
\centering
\includegraphics[width=.75\linewidth]{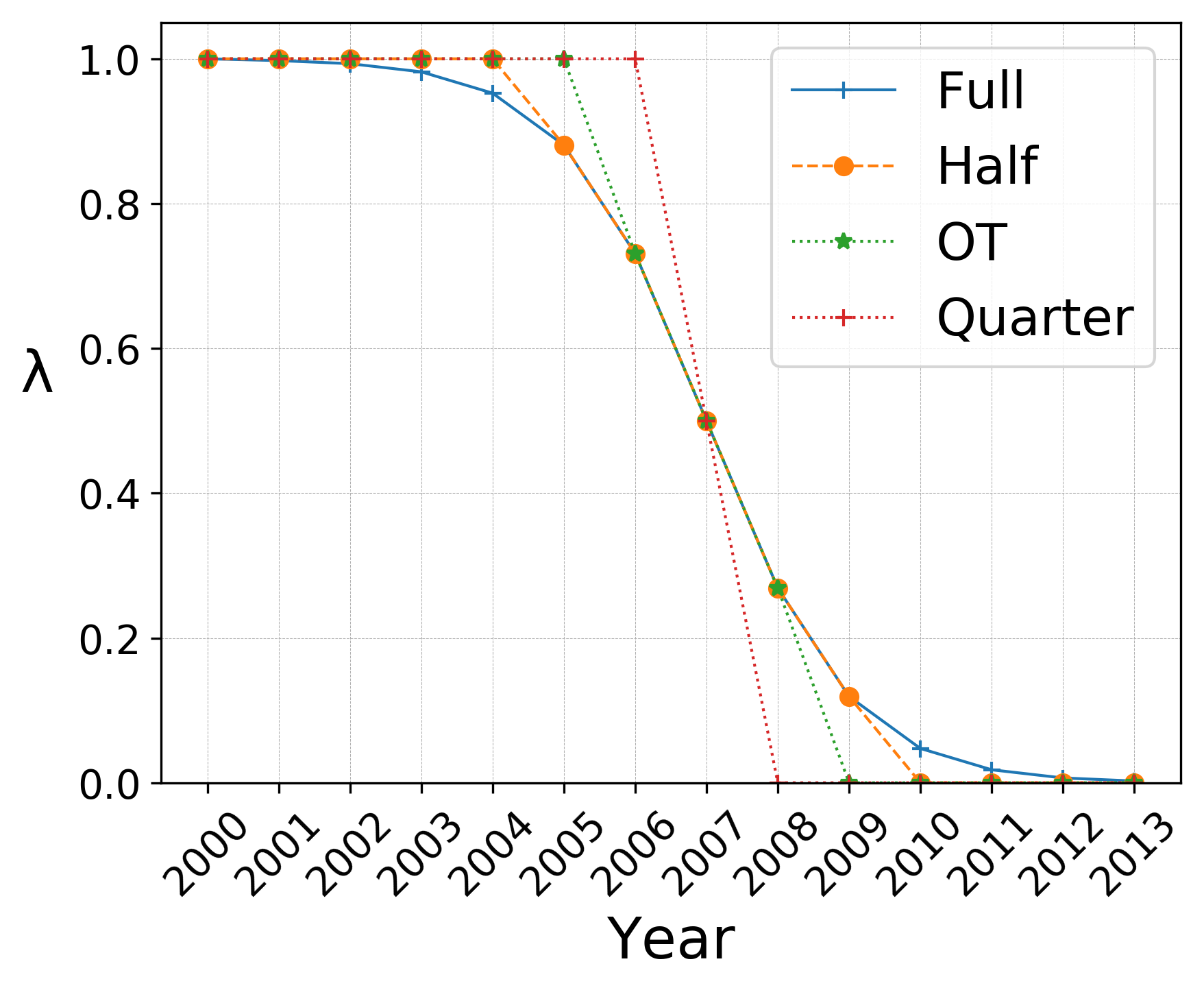}
\caption{The different functions used to model $\lambda_t$ in Eq.~\ref{eq:lambda}, indicating the speed and duration of the semantic change of our synthetic examples (see section~\ref{sec:conditioning2}).}
\label{fig:sigmoids}
\end{figure}

\vspace{.2cm}
\noindent\textbf{Conditioning on Duration of Change} \hspace{.15cm}
\label{sec:conditioning2}
The duration of the semantic change affects the value of $\lambda_t$ in Eq.~\ref{eq:lambda}. We conventionally set  $\lambda_{07}=0.5$, s.t. the artificial word representation  $w_{07}^{*(\alpha)}$ of a source word $\alpha$ in the year 2007 (i.e., the middle between 2001-2013) to be equal to $0.5 (w_{07}^{(\alpha)}  + w_{07}^{(\beta)}$). We then experiment with four different duration [start, end] ranges for the semantic change: (a) ``Full'' [2001-13], (b) ``Half'' [2005-10], (c) ``OT'' (One-Third) [2006-09] and (d) ``Quarter'' [2007-08].
%(a) the process starts occurring in year 2001 and ends in year 2013 (``Full''); (b) the shift starts in 2005 and ends in 2010 (``Half''); (c) the shift begins in 2006 and ends in 2009 (``one third''--``OT''); (d) the shift begins in 2007 and ends in 2008 (``Quarter''). 
A longer lasting semantic change duration implies a smoother transition of word $\alpha$ towards the meaning of word $\beta$, and vice versa (see Figure~\ref{fig:sigmoids}). By generating synthetic examples of varying lengths of semantic change duration we are able to measure the performance of the models under different conditions.

\subsection{Artificial Data Experiment} \label{sec:artificialtask}
Our task is to rank the words in the test set by means of their level of semantic change. We first train our three models on the training set and then we apply them on the test set. Finally, we measure the semantic change level of a word by means of the average cosine similarity between the predicted and actual word representations at each time step of the decoder. Model performance is assessed via rank-based metrics \cite{basile2018exploiting,tsakalidis2019mining,shoemark2019broom}.

\paragraph{Model Training}
The following is applicable to training of models for both the artificial and real-world data experiments.
We define and train our models as follows:
\begin{itemize}
    \item \texttt{seq2seq$_r$}: the autoencoder (section~\ref{sec:methods_past}) receives and reconstructs the full sequence of the word representations in the training set: $[W_{00}, ..., W_{13}]\xrightarrow{}[W_{00}^r, ..., W_{13}^r]$.
    \item \texttt{seq2seq$_f$}: the future prediction model (section~\ref{sec:methods_future}) receives the representation of the words in the training set in the year 2000 and learns to predict the rest of the sequence: $[W_{00}]\xrightarrow{}[W_{01}^f, ..., W_{13}^f]$.
    \item \texttt{seq2seq$_{rf}$}: the multi-task model (section~\ref{sec:methods_multi}) is fed with the first half of the sequence of the word representations in the training set and jointly learns to (a) reconstruct the input sequence and (b) predict the word representations in the future: $[W_{00}, ..., W_{06}]\xrightarrow{}\{[W_{00}^r, ..., W_{06}^r], [W_{07}^f, ..., W_{13}^f]\}$.
\end{itemize}
We vary the input in terms of number of time steps for \texttt{seq2seq$_r$} and \texttt{seq2seq$_f$} so that the decoder in each model operates on the maximum possible output sequence, thus exploiting the semantic change of the words over the whole time period (see section~\ref{sec:equivanelt}). \texttt{seq2seq$_{rf}$} is expected to be insensitive to the number of input time steps, therefore we conventionally set it to half of the overall sequence. We keep 25\% of our training set for validation purposes and train our models using the Adam optimiser \cite{kingma2014adam}. Parameter selection is performed based on 25 trials using the Tree of Parzen Estimators algorithm of the hyperopt module \cite{bergstra2013making}, by means of the maximum average (i.e., per time step) cosine similarity in the validation set.\footnote{For the complete list of parameters tested, refer to Appendix~\ref{app:A}.}

\begin{figure*}
\centering
\begin{subfigure}{.263\textwidth} 
  \centering
  \includegraphics[width=\textwidth]{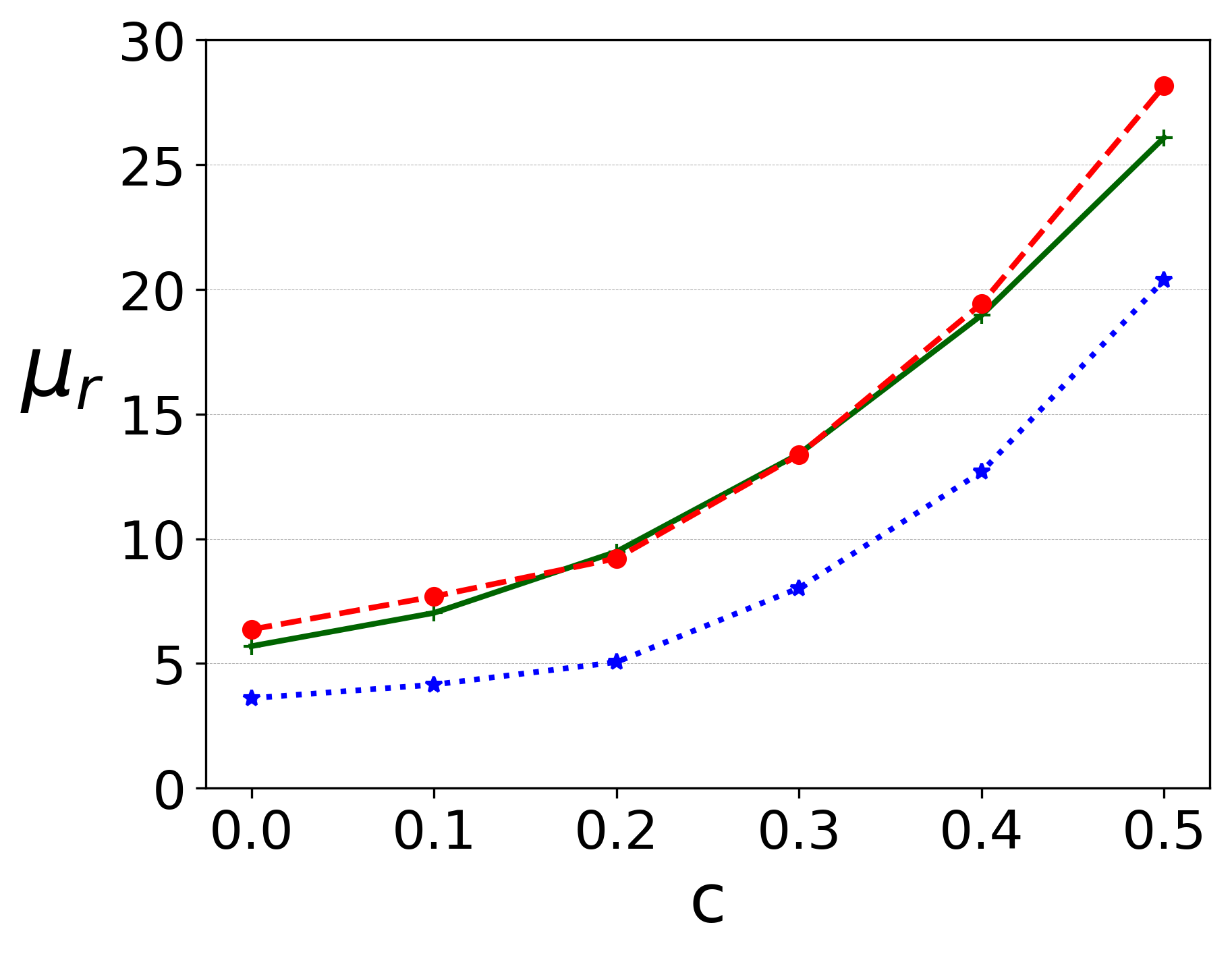}
  \caption{Full}
  \label{fig:sub11}
\end{subfigure}%
\begin{subfigure}{.24\textwidth}
  \centering
  \includegraphics[width=\textwidth]{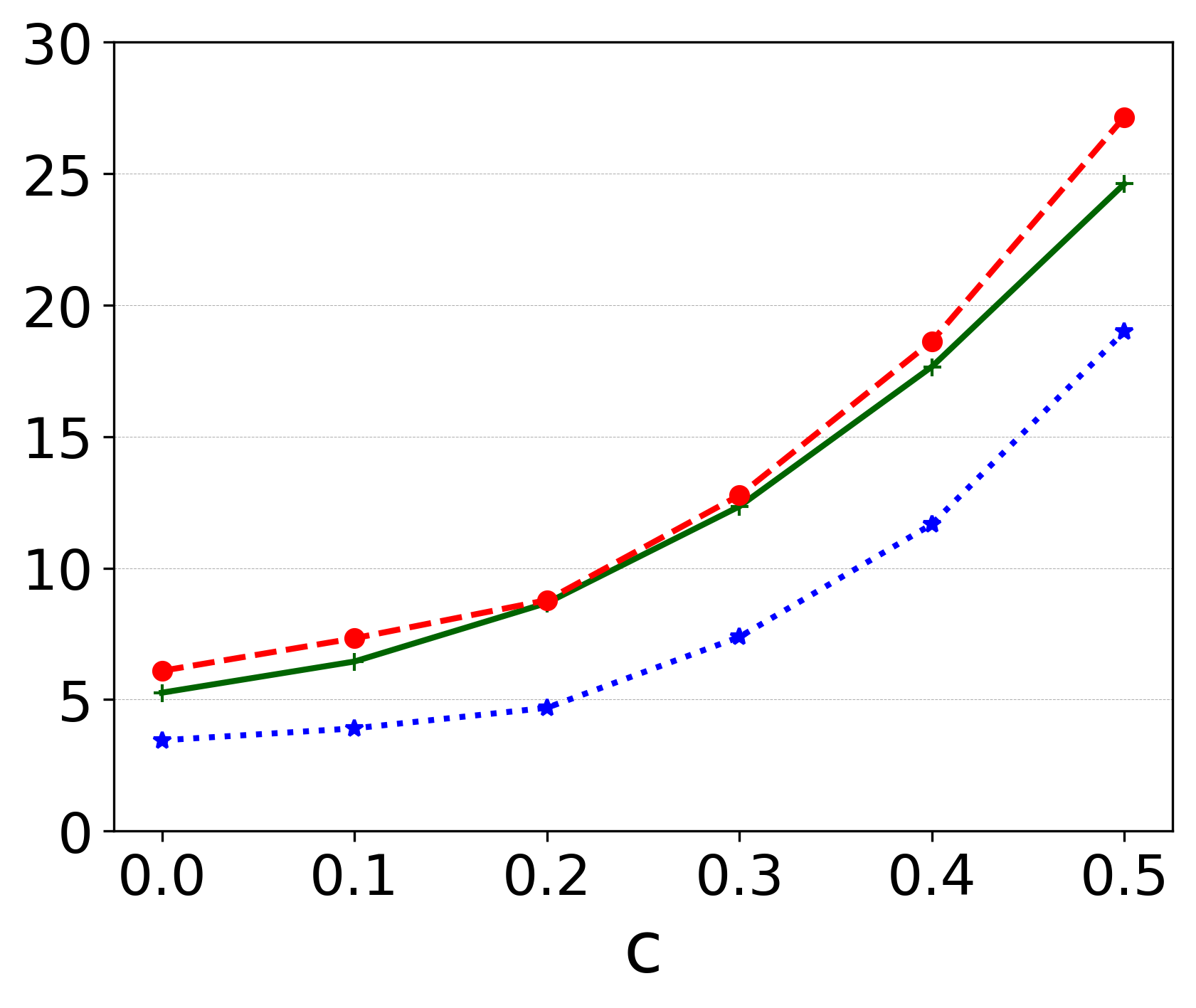}
  \caption{Half}
  \label{fig:sub12}
\end{subfigure}%
\begin{subfigure}{.24\textwidth}
  \centering
  \includegraphics[width=\textwidth]{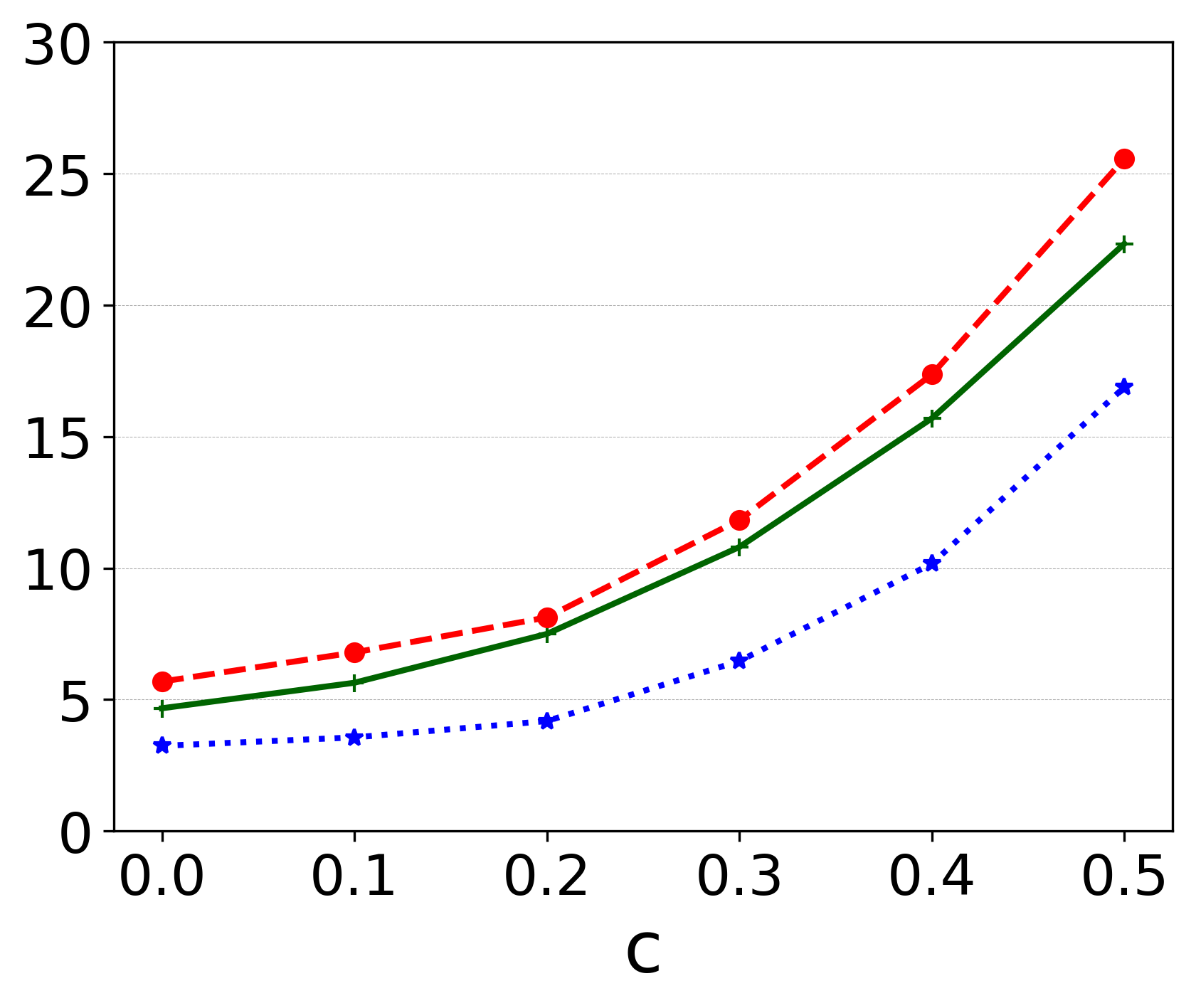}
  \caption{OT}
  \label{fig:sub12b}
\end{subfigure}%
\begin{subfigure}{.24\linewidth}
  \centering
  \includegraphics[width=\textwidth]{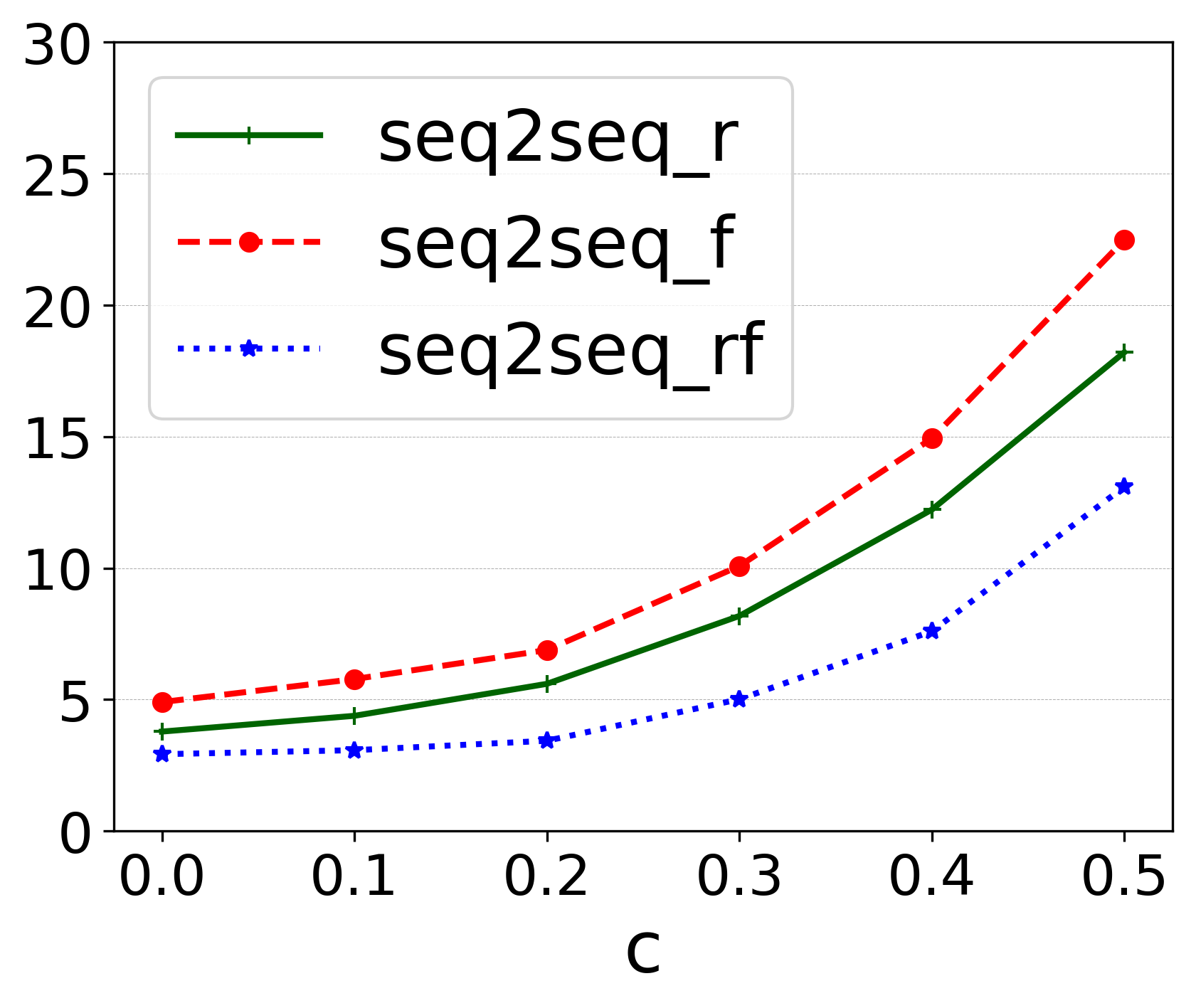}
  \caption{Quarter}
  \label{fig:sub13}
\end{subfigure}
\caption{$\mu_r$ of our models on the synthetic dataset for different values of the threshold $c$ and the four different periods of duration of semantic change (see \ref{sec:conditioning2}). Lower values of $\mu_r$ indicate a better performance.}
\label{fig:sigmoid_results}
\end{figure*}

\vspace{.18cm}
\noindent\textbf{Testing and Evaluation} \hspace{.15cm}
The following applies to experiments with both artificial and real-world data.
After training, each model is applied to the test set, yielding its predictions for every word across time.\footnote{Note that the future prediction model does not make a prediction for the first time step (year 2000).} The level of semantic change of a word in the test set is then calculated as the average cosine similarity between the actual and the predicted word representations through time \cite{hamilton2016diachronic,tsakalidis2019mining}, with higher values indicating a better model prediction -- thus, a lower level of semantic change. The words are ranked in descending order of their level of semantic change, so the lowest rank indicates a word whose vector representation has changed the most (i.e., indicating the most semantically shifted word).
For evaluation purposes, similarly to \citet{tsakalidis2019mining}, we employ the average rank across all of the semantically changed words (in \%, denoted as $\mu_r$), with lower scores indicating a better model. We prefer $\mu_r$ to the mean reciprocal rank, because the latter puts more weight on the first rankings. Since semantic change detection is an under-explored task in quantitative terms, we aim at getting better insights on model performance by working with an averaging metric such as $\mu_r$. For the same reason, in the current section we avoid using classification-based metrics that are based on a cut-off point (e.g., recall at $k$ \cite{basile2018exploiting}). We do make use of such metrics in the cross-model comparison in section~\ref{sec:results}.

\subsection{Results}\label{sec:synthetics_results}
%The results of the three models across all ($c$, $\lambda$) combinations are presented in Figure~\ref{fig:sigmoid_results}. Here, we outline the major findings with respect to cross-model comparison and the effect of the ($c$, $\lambda$) parameters in the performance of our models.

\vspace{.2cm}
\noindent\textbf{Model Comparison} \hspace{.15cm}  Figure~\ref{fig:sigmoid_results} presents the results of the three models on our synthetic data across all ($c$, $\lambda$) combinations. \texttt{seq2seq$_{rf}$} performs consistently better than the individual reconstruction (\texttt{seq2seq$_{r}$}) and future prediction (\texttt{seq2seq$_{f}$}) models across all experimental settings, showcasing that combining the two models under a multi-task setting benefits from the joint and finer-grained parameter tuning of the two components. The autoencoder performs slightly better than \texttt{seq2seq$_{f}$} -- a difference partially attributed to the fact that the autoencoder has a longer sequence to output ($W_{00}^r$), which helps explore the temporal variation of the words more effectively.

\begin{figure}
\centering
\begin{subfigure}{.16\textwidth}%32
  \centering
  \includegraphics[width=\textwidth]{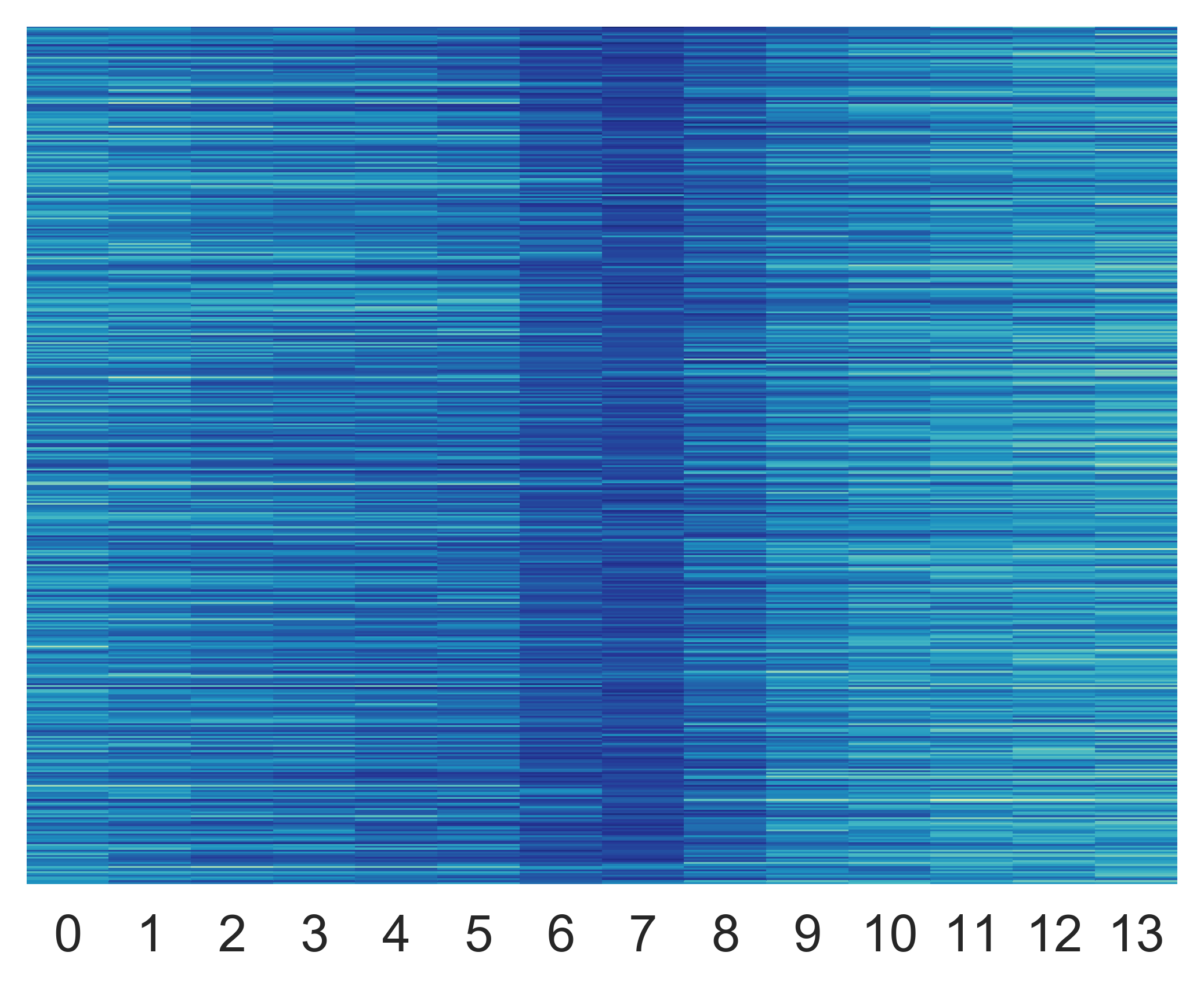}
  \caption{\texttt{seq2seq$_r$}}
  \label{fig:sub1}
\end{subfigure}%
\begin{subfigure}{.16\textwidth}%32
  \centering
  \includegraphics[width=\textwidth]{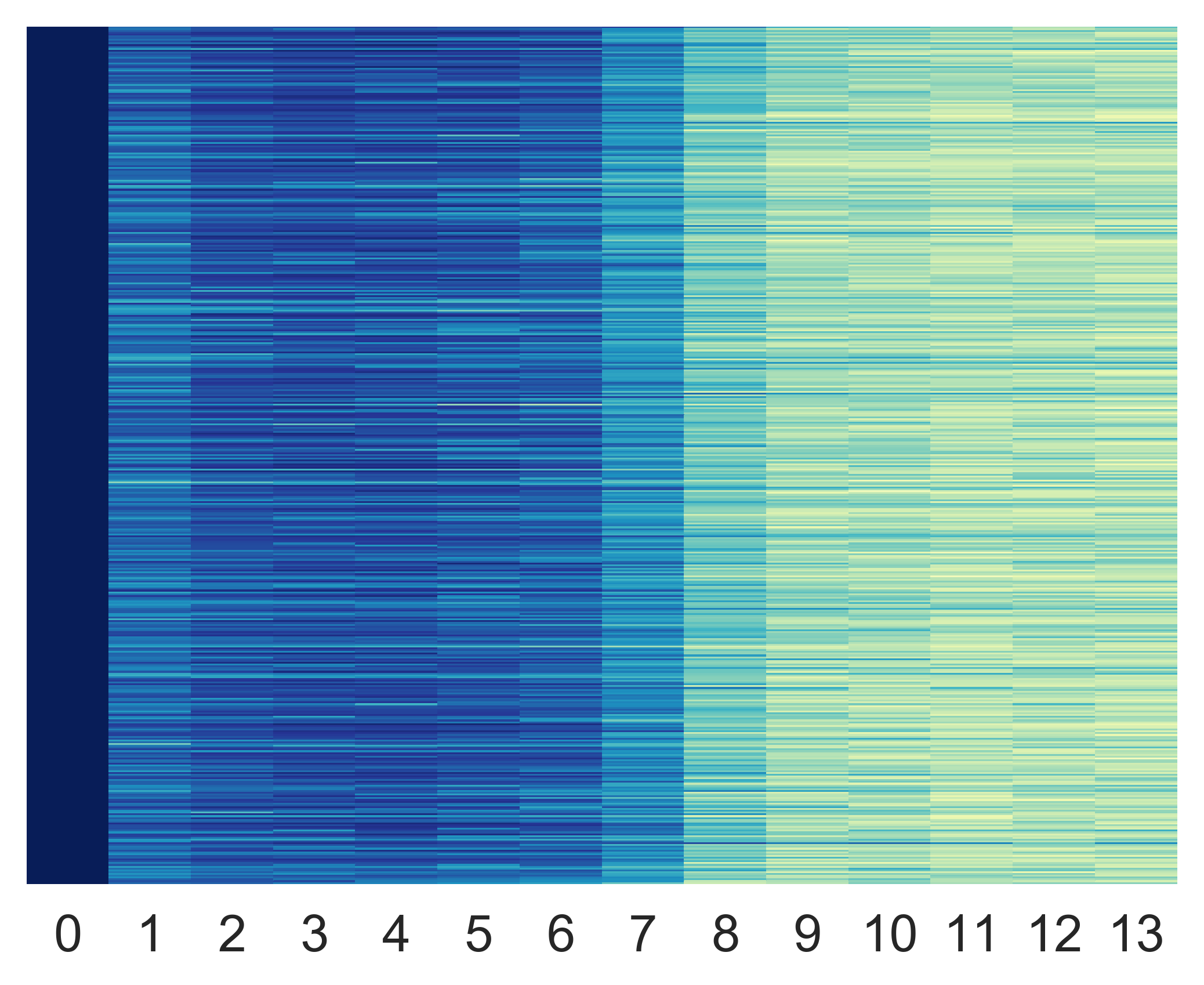}
  \caption{\texttt{seq2seq$_f$}}
  \label{fig:sub2}
\end{subfigure}%
\begin{subfigure}{.16\textwidth}%32
  \centering
  \includegraphics[width=\textwidth]{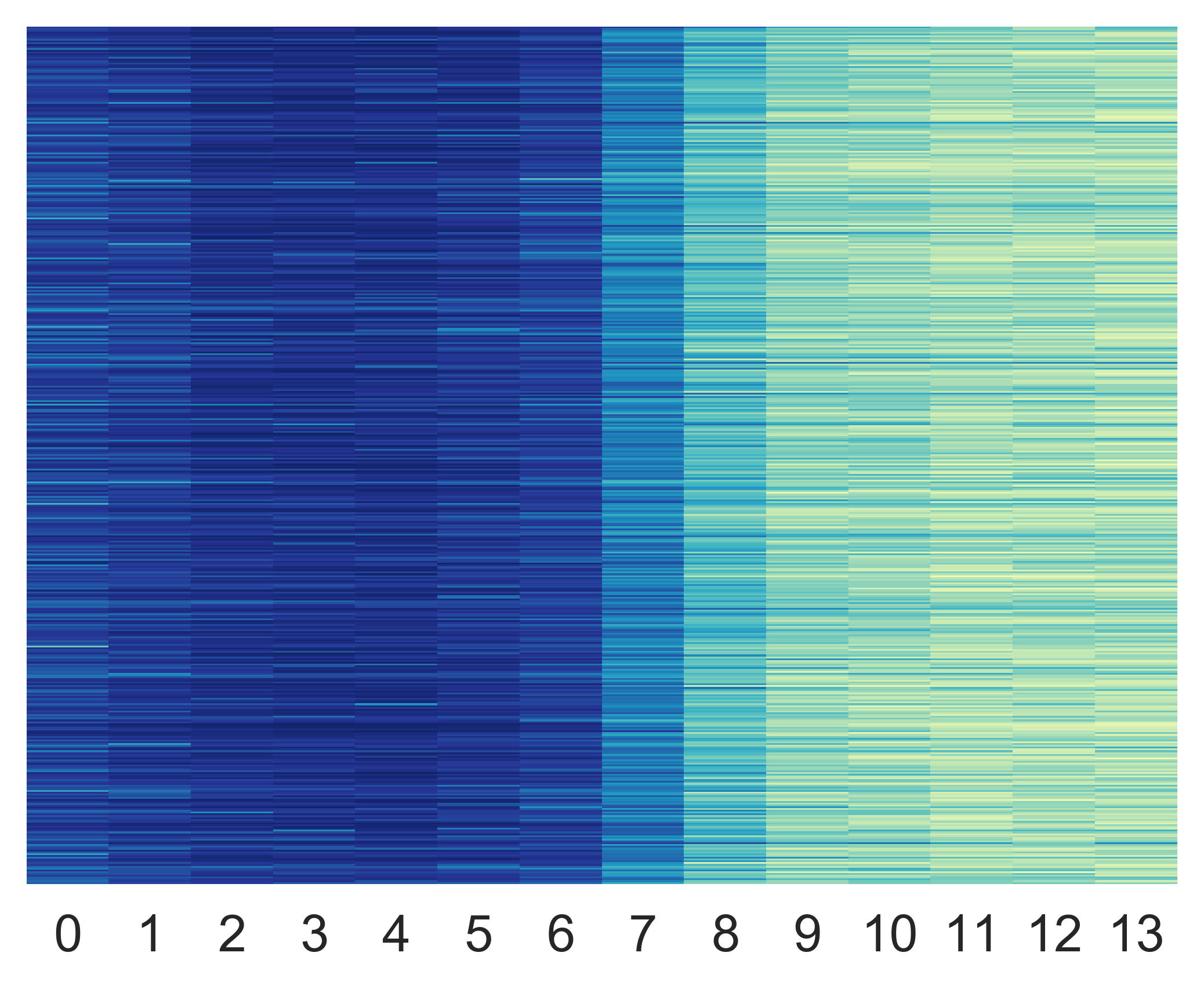}
  \caption{\texttt{seq2seq$_{rf}$}}
  \label{fig:sub3}
\end{subfigure}%
\begin{subfigure}{.02\textwidth}%02
  \centering
  \includegraphics[height=2.64cm,keepaspectratio]{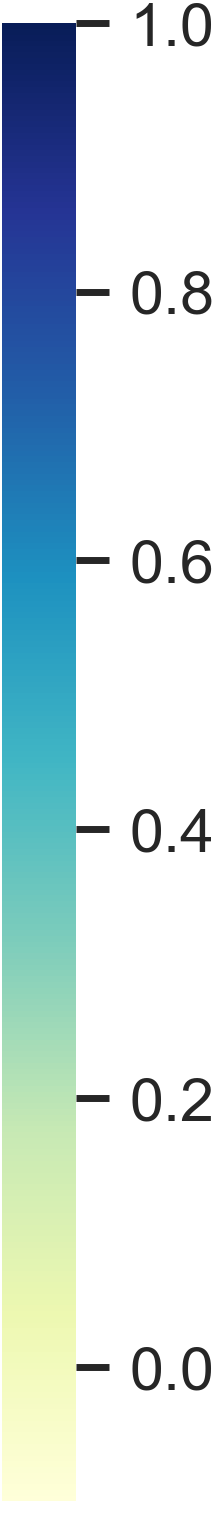} %4.68cm
\end{subfigure}
\caption{Cosine similarity between the actual and the predicted word vectors of the synthetic words that have undergone artificial semantic change (rows), per year (columns). Lighter colours indicate poorer model performance -- thus indicating that the corresponding words have undergone semantic change. Note that \texttt{seq2seq$_f$} does not make a prediction for the first time step (i.e., year 2000).}
\label{fig:cosines}
\end{figure}

Figure~\ref{fig:cosines} shows the cosine similarity between the predicted and actual representation of each synthetic word per time step for the ``Full'' case when $c$=0.0 (highest level of change, see section~\ref{sec:conditioning1}). A darker colour indicates a better model prediction -- thus a lower level of semantic change. \texttt{seq2seq$_{r}$} reconstructs the input sequence of the synthetic examples more accurately than the future prediction component (average cosine similarity per year ($avg\_cos$): .65 vs .50). It particularly manages to reconstruct the synthetic word representations \textit{during} the years 2006-2008 ($avg\_cos_{06:08}$=.75), which are the points when $\lambda_t$ varies more rapidly (see Figure~\ref{fig:sigmoids}); however, it fails to reconstruct equally well their representations before ($avg\_cos_{00:05}$= .65) and after ($avg\_cos_{09:13}$= .59) this sharp change. On the contrary, \texttt{seq2seq$_f$} predicts more accurately the synthetic word representations during the first years ($avg\_cos_{01:05}$ = .74), when the change in their semantics is minor, but completely fails  after the semantic change is almost complete (i.e., when $\lambda_t\leq.25$, $avg\_cos_{09:13}$= .24). \texttt{seq2seq$_{rf}$} benefits from the individual components' advantage: it appropriately reconstructs the artificial examples in the first years ($avg\_cos_{00:05}$ = .85) so that their semantic shift is highlighted more clearly during ($avg\_cos_{06:08}$= .62) and after the process is almost complete ($avg\_cos_{09:13}$= .26). Finally, $avg\_cos$ in \texttt{seq2seq$_{rf}$} highly correlates with $\lambda_t$ ($\rho$=.987), potentially providing insights on how to measure the speed of semantic change of a word. %We plan to investigate this effect in our future work.

\vspace{.2cm}
\noindent\textbf{Effect of Conditioning Parameters} \hspace{.15cm} 
 Regardless of the duration of the semantic change process and the model under consideration, an increase in the value of $c$ results in performance degradation. This is expected, since the increase of $c$ implies that the level of semantic change of the source words is lower, as discussed in \ref{sec:conditioning1}, thus making the task of detecting them more difficult. Nevertheless, our worst performing model in the most challenging setting ($c$=0.5, Full, seq2seq$_f$) achieves $\mu_r$=28.17, which is clearly better than the average $\mu_r$, expected by a random baseline ($\mu_r$=50.00). 

The decrease of the duration of semantic change has a positive effect on our models (see Figure~\ref{fig:sigmoid_results}). This is more evident in the cases of high value of $c$, where \texttt{seq2seq$_r$} ($\mu_r$: 26.09-18.21 in the Full-to-Quarter cases), \texttt{seq2seq$_f$} ($\mu_r$: 28.17-22.48) and \texttt{seq2seq$_{rf}$} ($\mu_r$:20.38-13.09) all show important gains in performance. This %is highly plausible, since it 
indicates that the models can capture the semantic change in small sub-sequences of the time-series. Studying this effect in datasets with a longer time span is an important future direction.

\section{Model Comparison with Real-World Data}\label{sec:experiments}

\begin{table*}[]
\centering
%\begin{minipage}[h]{0.6\textwidth}
\resizebox{.8\textwidth}{!}{%
\begin{tabular}{|c|l|cc|cc|cc|cc|}\hline
                  & & \multicolumn{2}{c|}{$\boldsymbol{\mu_r}$} & \multicolumn{2}{c|}{\textbf{Rec@5}} & \multicolumn{2}{c|}{\textbf{Rec@10}} & \multicolumn{2}{c|}{\textbf{Rec@50}}\\
                  & & '00-'13& avg$\pm$std & '00-'13& avg$\pm$std & '00-'13& avg$\pm$std & '00-'13 & avg$\pm$std  \\ \hline
   \multirow{10}{*}{\rotatebox[origin=c]{90}{Past Work/Baselines}}  & RAND& 49.97&  50.01$\pm$0.04&  5.00&  4.99$\pm$0.03&  10.01&  9.98$\pm$0.04&  50.02&  49.97$\pm$0.08\\\cline{2-10}
   &PROCR&            30.63&28.51$\pm$2.68&18.46&14.32$\pm$5.00&27.69&29.94$\pm$4.64&78.46&\textbf{80.47$\pm$3.79}\\
   & PROCR$_k$ & 31.47 &  28.71$\pm$2.65&  20.00& 14.67$\pm$3.85 &  29.23& 28.76$\pm$4.32 &  72.31&  79.64$\pm$4.49\\
   &PROCR$_{kt}$ &    31.91 &     28.47$\pm$2.85&  20.00& 14.32$\pm$4.23& 27.69& 28.88$\pm$4.45& 70.77& 80.00$\pm$4.53  \\\cline{2-10}
   
   &  RF &  30.01 &  30.45$\pm$4.15&  10.77&  15.62$\pm$4.30&  21.54&  27.46$\pm$7.16&  78.46&  77.63$\pm$6.42 \\
   &  LSTM$_r$&  27.87 &  \textbf{27.83$\pm$2.65} &  12.31 &  15.98$\pm$5.94 &  29.23 &  30.30$\pm$6.39 &  80.00 &  80.12$\pm$4.72 \\
   &  LSTM$_f$&  28.62 &  28.61$\pm$3.47 &  16.92 &  \textbf{17.40$\pm$5.60} &  32.31 &  \textbf{31.83$\pm$6.07} &  76.92 &  78.82$\pm$4.83\\\cline{2-10}
   
    &  GT$_c$&  47.87 &  44.04$\pm$1.54 &  7.69 & 7.41$\pm$2.26 & 16.92 &  14.13$\pm$3.76 & 52.31  &  57.90$\pm$2.94 \\

    &  GT$_\beta$& 38.09  &  36.16$\pm$1.74&  13.85 &  14.83$\pm$4.14&  24.62&  23.36$\pm$3.94&  66.15 &  69.37$\pm$3.26 \\ 
 
   & PROCR$_*$& 25.01 &  27.99$\pm$3.03 & 21.54 & 15.15 $\pm$4.52 & 32.31 &  28.40$\pm$3.75 & 81.54& 80.24$\pm$3.49
   \\\hline
  
   \multirow{3}{*}{\rotatebox[origin=c]{90}{Ours}} 
   &  seq2seq$_r$ &   24.75 &   28.36$\pm$3.38 &   21.54 &   19.05$\pm$4.47 &   38.46 &   29.94$\pm$6.64 &   \textbf{84.62} &   81.42$\pm$4.64 \\

   &seq2seq$_f$ & \textbf{23.86} & 27.17$\pm$4.16 & 26.15 & 22.01$\pm$6.72 & \textbf{46.15} & 34.32$\pm$10.13 & \textbf{84.62}& 81.18$\pm$5.07 \\

   &  seq2seq$_{rf}$ &   24.28 &   \textbf{24.29$\pm$0.67} &   \textbf{29.23}  &   \textbf{25.77$\pm$2.28} &   36.92 &   \textbf{39.49$\pm$2.11} &   \textbf{84.62} &   \textbf{85.00$\pm$1.16} \\\hline
\end{tabular}}%
\caption{Performance of our models and the baselines when operating on the entire time sequence (2000-2013) and averaged across time (2000-01, ..., 2000-13). \texttt{PROCR} and \texttt{PROCR$_{k(t)}$} are based on the methods employed in \citet{hamilton2016diachronic} and \citet{tsakalidis2019mining}, respectively; \texttt{GT$_{c,\beta}$} models are based on the work by \citet{shoemark2019broom}. The complete results in $\mu_r$ across all runs are provided in Appendix~\ref{app:B}.}
\label{tab:realresults}
%\end{minipage}\hfill
\end{table*}

\subsection{Experimental Setting}
We approach the task in a rank-based manner, as in section~\ref{sec:synthetic}. However, here we are interested in (a) detecting uncontrolled real-world examples of semantic change in words and (b) comparing our models against strong baselines and current practices.

\vspace{.2cm}
\noindent\textbf{Data and Task} \hspace{.15cm}
We make use of the UK Web Archive dataset (see section~\ref{sec:dataset1}). We keep the same 80/20 train/test split as in section~\ref{sec:synthetic} and incorporate in the test set the 65 words with known changes in meaning according to the Oxford English Dictionary. We train our models as in section~\ref{sec:artificialtask}, aiming at detecting (i.e., ranking lower) the 65 words in the test set. We use $\mu_r$ (as in section~\ref{sec:synthetic})and additionally recall at $k$ (Rec@k, k=5\%, 10\%, 50\%) as our evaluation metrics. Lower $\mu_r$ and higher Rec@k scores indicate better models.
\vspace{.15cm}

\vspace{.2cm}
\noindent\textbf{Models} \hspace{.15cm}
We compare the three variants from section~\ref{sec:methodology} %(\texttt{seq2seq$_r$}, \texttt{seq2seq$_f$}, \texttt{seq2seq$_{rf}$}) 
against four types of baselines:
\vspace{.1cm}

\noindent\textit{-- A random word rank generator} (\texttt{RAND}). We report average  metrics after 1K runs on the test set.
\vspace{.1cm}

\noindent\textit{-- Variants of Procrustes Alignment} \cite{schonemann1966generalized}, as the standard practice in past work \cite{hamilton2016diachronic,shoemark2019broom,tsakalidis2019mining}: Given the word representations in two different years $[W_0$, $W_i]$, \texttt{PROCR} transforms $W_i$ into $W^*_i$ s.t. the squared differences between $W_0$ and $W^*_i$ are minimised. We also use the \texttt{PROCR}$_k$ and \texttt{PROCR}$_{kt}$ variants \citep{tsakalidis2019mining}, which first detect the $k$ most stable words across either $[W_0$, $W_i]$ (\texttt{PROCR$_k$}) or $[W_0$, ..., $W_{T-1}]$ (\texttt{PROCR$_{kt}$}) to learn the alignment on and then transform $W_i$ into $W^*_i$. Words are ranked based on the cosine distance between $[W_0$, $W^*_i]$.
\vspace{.1cm}

\noindent\textit{-- Models leveraging the first and last word representations only}. We use a Random Forest \cite{breiman2001random} regression model (\texttt{RF}) that predicts $W_i$, given $W_0$. We also use the same architectures presented in sections~\ref{sec:methods_past}-\ref{sec:methods_future}, trained on $[W_0$, $W_i]$ (ignoring the full sequence): \texttt{LSTM$_r$} reconstructs the sequence $[W_0$, $W_i]$; \texttt{LSTM$_f$} predicts $W_i$, given $W_0$, similarly to \texttt{RF}. Words are ranked in inverse order of the (average, for \texttt{LSTM$_r$}) cosine similarity between their predicted and actual representations.

\noindent\textit{-- Models operating on the time series of distances}. Given a sequence of vectors $[W_0$, ..., $W_i]$, we construct the time series of cosine distances that result by \texttt{PROCR} \cite{kulkarni2015statistically,shoemark2019broom}. Then, we use two global trend models as in \citet{shoemark2019broom}: \texttt{GT$_c$} ranks the words by means of the absolute value of the Pearson correlation of their time series; \texttt{GT$_\beta$} fits instead a linear regression model for every word and ranks the words by the absolute value of the slope. Finally, we employ \texttt{PROCR$_*$}, ranking words based on the average cosine distance within $[0, i]$.\footnote{We refrain from evaluating the \texttt{GT} models when $i\leq$2, due to the very short time interval that does not allow for correlations to appear in the data, leading to very poor performance.}
\vspace{.1cm}

\iffalse
\begin{itemize}
    \item A random word rank generator (\texttt{RAND}). We perform 1,000 runs on the test set and report average performance metrics.
    
    \item Procrustes Alignment \cite{schonemann1966generalized}  and its variants, as the standard practice in related work \cite{hamilton2016diachronic,shoemark2019broom,tsakalidis2019mining}: Given the word representations in two different years \{$W_0$, $W_i$\}, \texttt{PROCR} transforms $W_i$ into $W^*_i$ s.t. the squared differences between $W_0$ and $W^*_i$ are minimised. We also employ the \texttt{PROCR}$_k$ and \texttt{PROCR}$_{kt}$ variants \citep{tsakalidis2019mining}, which first detect the $k$ most stable words -- across \{$W_0$, $W_i$\}, in the case of \texttt{PROCR$_k$}, or across \{$W_0$, $W_1$, ..., $W_{i-1}$, $W_i$\}, in \texttt{PROCR$_{kt}$} -- to learn the alignment on and then transform $W_i$ into $W^*_i$ based on them.
    
    \item Models leveraging the first and last word representations only. In specific, we use a Random Forest \cite{breiman2001random} regression model (\texttt{RF}) that predicts $W_i$, given $W_0$. We also use the same architectures presented in sections~\ref{sec:methods_past} and \ref{sec:methods_future}, trained on \{$W_0$, $W_i$\} only -- i.e., ignoring the full sequence: \texttt{LSTM$_r$} reconstructs the sequence \{$W_0$, $W_i$\}; \texttt{LSTM$_f$} predicts $W_i$, given $W_0$, similarly to \texttt{RF}.
    
    \item Models operating on the distances. We also employ \texttt{PROCR$_*$}
\end{itemize}
\fi

We report the performance of our models and baselines\footnote{All parameters tested during the training process of our baselines are provided in Appendix~\ref{app:A}.} (a) when they operate on the full interval [2000-2013] and (b) averaged across all intermediate intervals [2000-2001, ..., 2000-2013]. In the latter case, our models use additional (future) information compared to our baselines (e.g., when \texttt{seq2seq$_f$} is fed with the word sequences of [2000, 2001], it makes a prediction for the years [2002, ..., 2013] -- such information cannot be leveraged by the baselines). Thus, for (b), we only perform intra-model (and intra-baseline) comparisons.

\subsection{Results}\label{sec:results}
\vspace{.1cm}
\noindent\textbf{Our models \textit{vs} baselines} \hspace{.15cm}
The results are shown in Table~\ref{tab:realresults}. The three models proposed in this work consistently achieve the lowest $\mu_r$ and highest Rec@$k$ when working on the whole time sequence ('00-'13 columns in Table~\ref{tab:realresults}). The comparison between \{\texttt{seq2seq$_r$}, \texttt{LSTM$_r$}\} and \{\texttt{seq2seq$_f$}, \texttt{LSTM$_f$}\} in the years 2000-13 showcases the benefit of modelling the full sequence of the word representations across time, compared to using the first and last representations only. Overall, our models provide a relative boost of 4.6\% in $\mu_r$ and [35.7\%, 42.8\%, 5.8\%] in Rec@$k$ (for $k$=[5, 10, 50]) compared to the best performing baseline. \texttt{seq2seq$_f$} and \texttt{seq2seq$_{rf}$} models outperform the autoencoder (\texttt{seq2seq$_r$}) in most metrics, while \texttt{seq2seq$_{rf}$} yields the most stable results across all experiments. We explore these differences in detail in the last paragraph of this section.

\vspace{.2cm}
\noindent\textbf{Intra-baseline comparison} \hspace{.15cm}
Models operating only on the first and last word representations fail to outperform the simplistic Procrustes-based baselines in Rec@$k$, demonstrating again the weakness of operating in a non-sequential manner. The  \texttt{LSTM} models achieve low $\mu_r$ on the 2000-13 experiments; however, the difference with the rest of the baselines in $\mu_r$ across all years is negligible. The intra-Procrustes model comparison shows that the benefit of selecting a few anchor words to learn a better alignment (\texttt{PROCR$_k$}, \texttt{PROCR$_{kt}$}) shown in \citet{tsakalidis2019mining} in examining semantic change over two consecutive years does not apply when examining a longer time period. Finally, contrary to \citet{shoemark2019broom}, we find that time sensitive models operating on the word distances across time (\texttt{GT$_c$}, \texttt{GT$_\beta$}) perform worse than the baselines that leverage only the first and last word representations. This difference is attributed to the low number of time steps in our dataset that does not allow the \texttt{GT} models to exploit long-term correlations (i.e., considering the average distance across time (\texttt{PROCR$_*$}) performs better), but also highlights the importance of leveraging the full word sequence across time.
%-- i.e., \texttt{PROCR$_k$} and \texttt{PROCR$_{kt}$} fail to confidently outperform \texttt{PROCR} in all of our evaluation metrics.

\includegraphics[width=0.75\linewidth]{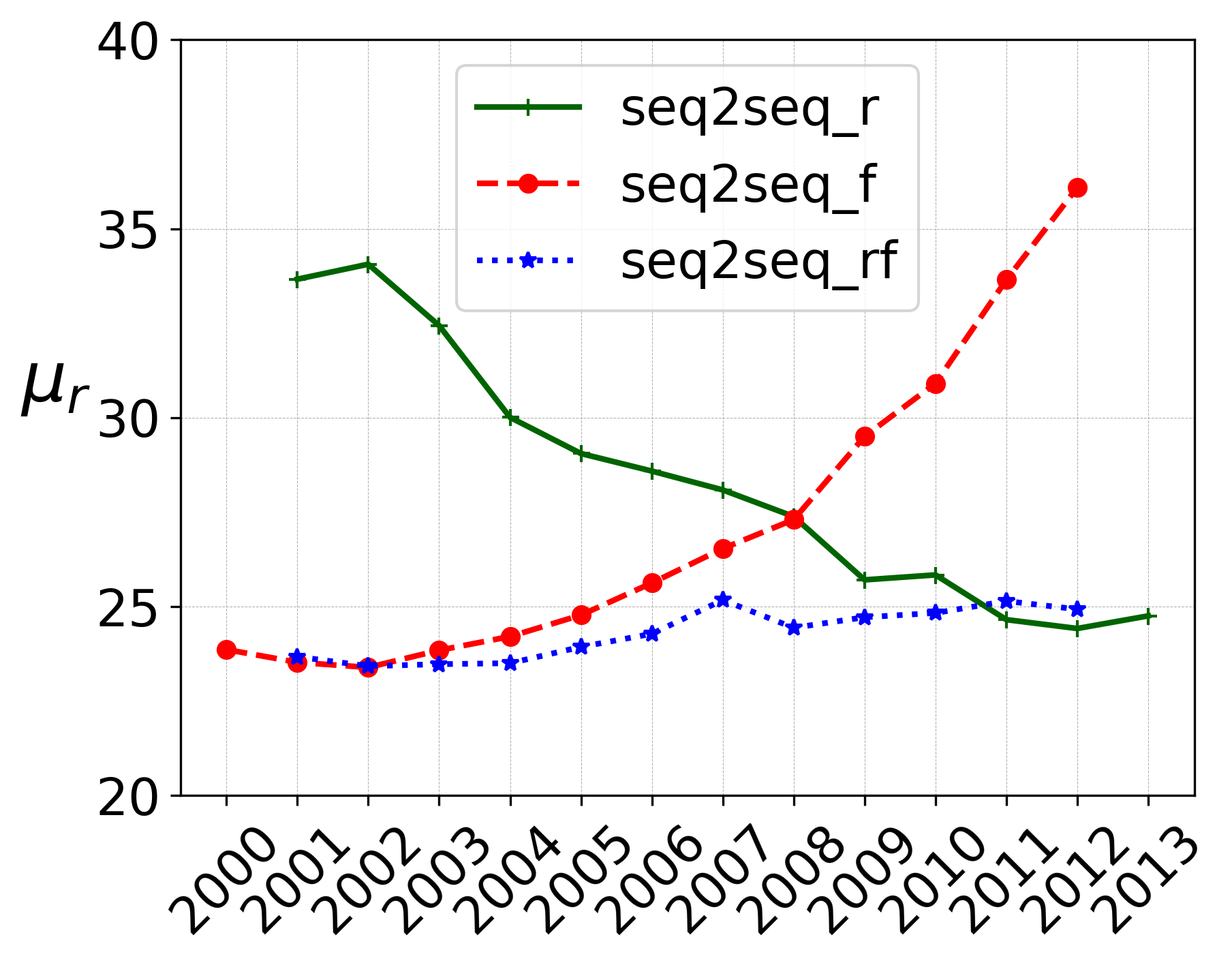}
\captionof{figure}{$\mu_r$ of our models for varying value of $i$ (Eq.~\ref{eq:past}--\ref{eq:multi}).\footnotemark}
\label{fig:final_chart}
\vspace{.2cm}

\noindent\textbf{Effect of input/output lengths}\hspace{.15cm}
Figure~\ref{fig:final_chart} shows the $\mu_r$ of our three variants when we alter the length of the input and, therefore, also the length of the output (see section~\ref{sec:equivanelt}). \footnotetext{Example: For the year 2005 (x-axis), all models receive the word representations until 2005 as their input. Then, \texttt{seq2seq$_r$} reconstructs the word representations up to 2005, \texttt{seq2seq$_f$} predicts the future representations (2006, ..., 2013) and \texttt{seq2seq$_{rf}$} performs both tasks jointly.} The performance of  \texttt{seq2seq$_r$} increases with the input size since by definition the decoder is able to detect words whose semantics have changed over a longer period of time (i.e., within $[2000, i]$, with $i$ increasing), while also modelling a longer sequence of a word's representation through time. On the contrary, the performance of \texttt{seq2seq$_f$} increases alongside the decrease of the number of input time steps. This is expected since, as $i$ decreases, \texttt{seq2seq$_f$} encodes a shorter input sequence and the decoding (and hence the semantic change detection) is applied on the remaining (and increased number of) time steps within $[i+1, 2013]$. These findings provide empirical evidence that both models can achieve better performance if trained over longer sequences of time steps. Finally, the stability of \texttt{seq2seq$_{rf}$} showcases its input length-invariant nature, which is also clearly evident in all of the averaged results (standard deviation in avg$\pm$std columns) in Table~\ref{tab:realresults}: in its worst performing setting, \texttt{seq2seq$_{rf}$} still manages to achieve results that are close to the best performing model ($\mu_r$=25.17, Rec@$k$=[21.54, 36.92, 83.08] for the three thresholds) and always better (or equal to) the best performing baseline shown in Table~\ref{tab:realresults} in Rec@$k$. This is %a rather plausible property, since 
 a very attractive aspect of the model as it removes the need to manually define the number of time steps to be fed to the encoder.  %it releases us from manually defining the number of time steps that are fed to the encoder, and thus the overall architecture with respect to the number of time steps being processed at each stage of the model.

\vspace{.2cm}

\section{Conclusion and Future Work}
We have proposed three variants of %seq2seq
sequential models for semantic change detection that effectively exploit the full sequence of a word's representation through time to determine its level of semantic change. 
%This is done by measuring the error that results from (a) predicting the future representations of the word, (b) reconstructing the word representations through time, or (c) jointly performing both tasks. 
Through extensive experimentation based on synthetic and real-world data, we have demonstrated that the proposed models can %successfully tackle the task, 
surpass state-of-the-art results on the UK Web Archive Dataset. Importantly, their performance increases alongside the duration of the time period under study, confidently outperforming competitive baselines and common practices in the literature on semantic change. %By releasing our resources, we aim at providing a basis for model comparison in semantic change detection. 

In future work we plan to incorporate anomaly detection approaches operating on the model's predicted word vectors instead of considering the average similarity between the predicted and the actual representations as the level of semantic change of a word. Employing contextual word representations \cite{devlin2019bert,hu2019diachronic} can also be of high importance in detecting new senses of the words across time. Finally, we plan to investigate different architectures, such as Variational Autoencoders \cite{kingma2013auto}, and test our models in datasets of different duration and in different languages to provide clearer evidence on their effectiveness.

\section*{Acknowledgements}
This work was supported by The Alan Turing Institute (grant EP/N510129/1) and by a Turing AI Fellowship to Maria Liakata, funded by the Department of Business, Energy \& Industrial Strategy.

\bibliography{acl2020}

\begin{thebibliography}{34}
\expandafter\ifx\csname natexlab\endcsname\relax\def\natexlab#1{#1}\fi

\bibitem[{Bamler and Mandt(2017)}]{bamler2017dynamic}
Robert Bamler and Stephan Mandt. 2017.
\newblock Dynamic {W}ord {E}mbeddings.
\newblock In \emph{Proceedings of the 34th International Conference on Machine
  Learning-Volume 70}, pages 380--389. JMLR. org.

\bibitem[{Basile and McGillivray(2018)}]{basile2018exploiting}
Pierpaolo Basile and Barbara McGillivray. 2018.
\newblock Exploiting the {W}eb for {S}emantic {C}hange {D}etection.
\newblock In \emph{International Conference on Discovery Science}, pages
  194--208. Springer.

\bibitem[{Bergstra et~al.(2013)Bergstra, Yamins, and Cox}]{bergstra2013making}
James Bergstra, Daniel Yamins, and David~Daniel Cox. 2013.
\newblock Making a {S}cience of {M}odel {S}earch: {H}yperparameter
  {O}ptimization in {H}undreds of {D}imensions for {V}ision {A}rchitectures.
\newblock pages 115--123.

\bibitem[{Breiman(2001)}]{breiman2001random}
Leo Breiman. 2001.
\newblock Random {F}orests.
\newblock \emph{Machine Learning}, 45(1):5--32.

\bibitem[{Cook and Stevenson(2010)}]{cook2010automatically}
Paul Cook and Suzanne Stevenson. 2010.
\newblock Automatically {I}dentifying {C}hanges in the {S}emantic {O}rientation
  of {W}ords.
\newblock In \emph{Proceedings of the Seventh conference on International
  Language Resources and Evaluation}.

\bibitem[{Danescu-Niculescu-Mizil et~al.(2013)Danescu-Niculescu-Mizil, West,
  Jurafsky, Leskovec, and Potts}]{Leskovec13a}
Cristian Danescu-Niculescu-Mizil, Robert West, Dan Jurafsky, Jure Leskovec, and
  Christopher Potts. 2013.
\newblock No {C}ountry for {O}ld {M}embers: {U}ser {L}ifecycle and {L}inguistic
  {C}hange in {O}nline {C}ommunities.
\newblock In \emph{Proceedings of the 22nd International Conference on World
  Wide Web}, pages 307--318. Association for Computing Machinery.

\bibitem[{Del~Tredici et~al.(2019)Del~Tredici, Fern{\'a}ndez, and
  Boleda}]{del2019short}
Marco Del~Tredici, Raquel Fern{\'a}ndez, and Gemma Boleda. 2019.
\newblock Short-{T}erm {M}eaning {S}hift: {A} {D}istributional {E}xploration.
\newblock In \emph{Proceedings of the 2019 Conference of the North American
  Chapter of the Association for Computational Linguistics: Human Language
  Technologies, Volume 1 (Long and Short Papers)}, pages 2069--2075.

\bibitem[{Devlin et~al.(2019)Devlin, Chang, Lee, and
  Toutanova}]{devlin2019bert}
Jacob Devlin, Ming-Wei Chang, Kenton Lee, and Kristina Toutanova. 2019.
\newblock {BERT}: {P}re-training of {D}eep {B}idirectional {T}ransformers for
  {L}anguage {U}nderstanding.
\newblock In \emph{Proceedings of the 2019 Conference of the North American
  Chapter of the Association for Computational Linguistics: Human Language
  Technologies, Volume 1 (Long and Short Papers)}, pages 4171--4186.

\bibitem[{Dubossarsky et~al.(2019)Dubossarsky, Hengchen, Tahmasebi, and
  Schlechtweg}]{dubossarsky2019time}
Haim Dubossarsky, Simon Hengchen, Nina Tahmasebi, and Dominik Schlechtweg.
  2019.
\newblock Time-{O}ut: {T}emporal {R}eferencing for {R}obust {M}odeling of
  {L}exical {S}emantic {C}hange.
\newblock In \emph{Proceedings of the 57th Annual Meeting of the Association
  for Computational Linguistics}, pages 457--470.

\bibitem[{Gulordava and Baroni(2011)}]{gulordava2011distributional}
Kristina Gulordava and Marco Baroni. 2011.
\newblock A distributional similarity approach to the detection of semantic
  change in the {G}oogle {B}ooks {N}gram corpus.
\newblock In \emph{Proceedings of the GEMS 2011 Workshop on Geometrical Models
  of Natural Language Semantics}, pages 67--71.

\bibitem[{Hamilton et~al.(2016)Hamilton, Leskovec, and
  Jurafsky}]{hamilton2016diachronic}
William~L Hamilton, Jure Leskovec, and Dan Jurafsky. 2016.
\newblock Diachronic {W}ord {E}mbeddings {R}eveal {S}tatistical {L}aws of
  {S}emantic {C}hange.
\newblock In \emph{Proceedings of the 54th Annual Meeting of the Association
  for Computational Linguistics (Volume 1: Long Papers)}, volume~1, pages
  1489--1501.

\bibitem[{Hochreiter and Schmidhuber(1997)}]{hochreiter1997long}
Sepp Hochreiter and J{\"u}rgen Schmidhuber. 1997.
\newblock Long {S}hort-{T}erm {M}emory.
\newblock \emph{Neural Computation}, 9(8):1735--1780.

\bibitem[{Hu et~al.(2019)Hu, Li, and Liang}]{hu2019diachronic}
Renfen Hu, Shen Li, and Shichen Liang. 2019.
\newblock Diachronic {S}ense {M}odeling with {D}eep {C}ontextualized {W}ord
  {E}mbeddings: {A}n {E}cological {V}iew.
\newblock In \emph{Proceedings of the 57th Annual Meeting of the Association
  for Computational Linguistics}, pages 3899--3908.

\bibitem[{Kim et~al.(2014)Kim, Chiu, Hanaki, Hegde, and
  Petrov}]{kim2014temporal}
Yoon Kim, Yi-I Chiu, Kentaro Hanaki, Darshan Hegde, and Slav Petrov. 2014.
\newblock Temporal {A}nalysis of {L}anguage through {N}eural {L}anguage
  {M}odels.
\newblock In \emph{Proceedings of the ACL 2014 Workshop on Language
  Technologies and Computational Social Science}, pages 61--65.

\bibitem[{Kingma and Ba(2015)}]{kingma2014adam}
Diederik~P. Kingma and Jimmy Ba. 2015.
\newblock Adam: {A} {M}ethod for {S}tochastic {O}ptimization.
\newblock In \emph{3rd International Conference on Learning Representations,
  {ICLR} 2015, Conference Track Proceedings}.

\bibitem[{Kingma and Welling(2014)}]{kingma2013auto}
Diederik~P. Kingma and Max Welling. 2014.
\newblock Auto-encoding variational bayes.
\newblock In \emph{2nd International Conference on Learning Representations,
  {ICLR} 2014, Conference Track Proceedings}.

\bibitem[{Kulkarni et~al.(2015)Kulkarni, Al-Rfou, Perozzi, and
  Skiena}]{kulkarni2015statistically}
Vivek Kulkarni, Rami Al-Rfou, Bryan Perozzi, and Steven Skiena. 2015.
\newblock Statistically significant detection of linguistic change.
\newblock In \emph{Proceedings of the 24th International Conference on World
  Wide Web}, pages 625--635. International World Wide Web Conferences Steering
  Committee.

\bibitem[{Kutuzov et~al.(2018)Kutuzov, {\O}vrelid, Szymanski, and
  Velldal}]{kutuzov2018diachronic}
Andrey Kutuzov, Lilja {\O}vrelid, Terrence Szymanski, and Erik Velldal. 2018.
\newblock Diachronic {W}ord {E}mbeddings and {S}emantic {S}hifts: {A} {S}urvey.
\newblock In \emph{Proceedings of the 27th International Conference on
  Computational Linguistics}, pages 1384--1397.

\bibitem[{McAuley and Leskovec(2013)}]{Leskovec13b}
Julian~John McAuley and Jure Leskovec. 2013.
\newblock From {A}mateurs to {C}onnoisseurs: {M}odeling the {E}volution of
  {U}ser {E}xpertise through {O}nline {R}eviews.
\newblock In \emph{Proceedings of the 22nd International Conference on World
  Wide Web}, pages 897--908. Association for Computing Machinery.

\bibitem[{Michel et~al.(2011)Michel, Shen, Aiden, Veres, Gray, Pickett,
  Hoiberg, Clancy, Norvig, Orwant et~al.}]{michel2011quantitative}
Jean-Baptiste Michel, Yuan~Kui Shen, Aviva~Presser Aiden, Adrian Veres,
  Matthew~K Gray, Joseph~P Pickett, Dale Hoiberg, Dan Clancy, Peter Norvig, Jon
  Orwant, et~al. 2011.
\newblock Quantitative {A}nalysis of {C}ulture {U}sing {M}illions of
  {D}igitized {B}ooks.
\newblock \emph{Science}, 331(6014):176--182.

\bibitem[{Mihalcea and Nastase(2012)}]{mihalcea2012word}
Rada Mihalcea and Vivi Nastase. 2012.
\newblock Word {E}poch {D}isambiguation: {F}inding how {W}ords {C}hange over
  {T}ime.
\newblock In \emph{Proceedings of the 50th Annual Meeting of the Association
  for Computational Linguistics (Volume 2: Short Papers)}, pages 259--263.

\bibitem[{Mikolov et~al.(2013)Mikolov, Sutskever, Chen, Corrado, and
  Dean}]{mikolov2013distributed}
Tomas Mikolov, Ilya Sutskever, Kai Chen, Greg~S Corrado, and Jeff Dean. 2013.
\newblock Distributed {R}epresentations of {W}ords and {P}hrases and their
  {C}ompositionality.
\newblock In \emph{Advances in Neural Information Processing Systems}, pages
  3111--3119.

\bibitem[{Rosenfeld and Erk(2018)}]{rosenfeld2018deep}
Alex Rosenfeld and Katrin Erk. 2018.
\newblock Deep {N}eural {M}odels of {S}emantic {S}hift.
\newblock In \emph{Proceedings of the 2018 Conference of the North American
  Chapter of the Association for Computational Linguistics: Human Language
  Technologies, Volume 1 (Long Papers)}, pages 474--484.

\bibitem[{Rudolph and Blei(2018)}]{rudolph2018dynamic}
Maja Rudolph and David Blei. 2018.
\newblock Dynamic {E}mbeddings for {L}anguage {E}volution.
\newblock In \emph{Proceedings of the 2018 World Wide Web Conference on World
  Wide Web}, pages 1003--1011. International World Wide Web Conferences
  Steering Committee.

\bibitem[{Sagi et~al.(2009)Sagi, Kaufmann, and Clark}]{sagi2009semantic}
Eyal Sagi, Stefan Kaufmann, and Brady Clark. 2009.
\newblock Semantic {D}ensity {A}nalysis: {C}omparing {W}ord {M}eaning across
  {T}ime and {P}honetic {S}pace.
\newblock In \emph{Proceedings of the Workshop on Geometrical Models of Natural
  Language Semantics}, pages 104--111. Association for Computational
  Linguistics.

\bibitem[{Schlechtweg et~al.(2019)Schlechtweg, H{\"a}tty, del Tredici, and
  Walde}]{schlechtweg2019wind}
Dominik Schlechtweg, Anna H{\"a}tty, Marco del Tredici, and Sabine Schulte~im
  Walde. 2019.
\newblock A {W}ind of {C}hange: {D}etecting and {E}valuating {L}exical
  {S}emantic {C}hange across {T}imes and {D}omains.
\newblock \emph{arXiv preprint arXiv:1906.02979}.

\bibitem[{Schlechtweg et~al.(2020)Schlechtweg, McGillivray, Hengchen,
  Dubossarsky, and Tahmasebi}]{schlechtweg2020semeval}
Dominik Schlechtweg, Barbara McGillivray, Simon Hengchen, Haim Dubossarsky, and
  Nina Tahmasebi. 2020.
\newblock {S}em{E}val-2020 {T}ask 1: {U}nsupervised {L}exical {S}emantic
  {C}hange {D}etection.
\newblock In \emph{To appear in Proceedings of the 14th International Workshop
  on Semantic Evaluation}, Barcelona, Spain. Association for Computational
  Linguistics.

\bibitem[{Schlechtweg et~al.(2018)Schlechtweg, im~Walde, and
  Eckmann}]{schlechtweg2018diachronic}
Dominik Schlechtweg, Sabine~Schulte im~Walde, and Stefanie Eckmann. 2018.
\newblock Diachronic {U}sage {R}elatedness ({DUR}el): {A} {F}ramework for the
  {A}nnotation of {L}exical {S}emantic {C}hange.
\newblock In \emph{Proceedings of the 2018 Conference of the North American
  Chapter of the Association for Computational Linguistics: Human Language
  Technologies, Volume 2 (Short Papers)}, pages 169--174.

\bibitem[{Sch{\"o}nemann(1966)}]{schonemann1966generalized}
Peter~H Sch{\"o}nemann. 1966.
\newblock A {G}eneralized {S}olution of the {O}rthogonal {P}rocrustes
  {P}roblem.
\newblock \emph{Psychometrika}, 31(1):1--10.

\bibitem[{Shoemark et~al.(2019)Shoemark, Ferdousi~Liza, Nguyen, Hale, and
  McGillivray}]{shoemark2019broom}
Philippa Shoemark, Farhana Ferdousi~Liza, Dong Nguyen, Scott~A Hale, and
  Barbara McGillivray. 2019.
\newblock Room to {G}lo: {A} {S}ystematic {C}omparison of {S}emantic {C}hange
  {D}etection {A}pproaches with {W}ord {E}mbeddings.
\newblock In \emph{Proceedings of the 2019 Conference on Empirical Methods in
  Natural Language Processing and the 9th International Joint Conference on
  Natural Language Processing}, pages 66--76.

\bibitem[{Srivastava et~al.(2014)Srivastava, Hinton, Krizhevsky, Sutskever, and
  Salakhutdinov}]{srivastava2014dropout}
Nitish Srivastava, Geoffrey Hinton, Alex Krizhevsky, Ilya Sutskever, and Ruslan
  Salakhutdinov. 2014.
\newblock Dropout: {A} {S}imple {W}ay to {P}revent {N}eural {N}etworks from
  {O}verfitting.
\newblock \emph{The Journal of Machine Learning Research}, 15(1):1929--1958.

\bibitem[{Tang(2018)}]{tang2018state}
Xuri Tang. 2018.
\newblock A {S}tate-of-the-{A}rt of {S}emantic {C}hange {C}omputation.
\newblock \emph{Natural Language Engineering}, 24(5):649--676.

\bibitem[{Tsakalidis et~al.(2019)Tsakalidis, Bazzi, Cucuringu, Basile, and
  McGillivray}]{tsakalidis2019mining}
Adam Tsakalidis, Marya Bazzi, Mihai Cucuringu, Pierpaolo Basile, and Barbara
  McGillivray. 2019.
\newblock Mining the {UK} {W}eb {A}rchive for {S}emantic {C}hange {D}etection.
\newblock In \emph{Recent Advances in Natural Language Processing}.

\bibitem[{Yao et~al.(2018)Yao, Sun, Ding, Rao, and Xiong}]{yao2018dynamic}
Zijun Yao, Yifan Sun, Weicong Ding, Nikhil Rao, and Hui Xiong. 2018.
\newblock Dynamic {W}ord {E}mbeddings for {E}volving {S}emantic {D}iscovery.
\newblock In \emph{Proceedings of the Eleventh ACM International Conference on
  Web Search and Data Mining}, pages 673--681. ACM.

\end{thebibliography}
\bibliographystyle{acl_natbib}

\appendix

\section{List of Hyperparameters}\label{app:A}
\paragraph{Our models} We test the following hyper-parameters for our \texttt{seq2seq$_{r/f/rf}$} models:
\begin{itemize}
    \item encoder\_LSTM$_0$, number of units: [32, 64, 128, 256, 512]
    \item encoder\_LSTM$_1$, number of units: [32, 64]
    \item decoder\_LSTM$_0$, number of units: [32, 64] (x2, for the case of \texttt{seq2seq$_{rf}$} -- for (a) the autoencoding and (b) future prediction component)
    \item decoder\_LSTM$_1$, number of units: [32, 64, 128, 256, 512] (x2, for the case of \texttt{seq2seq$_{rf}$})
    \item dropout rate in dropout layers: [.1, .25, .5]
    \item batch size: [32, 64, 128, 256, 512, 1024]
    \item number of epochs: [10, 20, 30, 40, 50]
\end{itemize}
We optimise our parameters using the Adam optimiser in \texttt{keras}, using the default learning rate (.001).  

\paragraph{Baselines}
We experiment with the following hyper-parameters per model:
\begin{itemize}
    \item \texttt{LSTM$_{r/f}$}: we follow the exact same settings as in our models.
    \item \texttt{RF}: we experiment with the number of trees ([50, 100, 150, 200]) and select the best model based on the maximum average cosine similarity across all predictions, as in our models. 
    \item \texttt{PROCR$_{k/kt}$}: we experiment with different rate [.001, .01, .05, .1, .2, ... .9] of anchor (or diachronic anchor) words on the basis of the size of the test set. We select to display in our results the best model based on the average performance in the test set ($k$=.9 for \texttt{PROCR$_{k}$}, $k$=.5 for \texttt{PROCR$_{kt}$}).
    \item \texttt{GT$_c$}: we explore different correlation metrics (Spearman Rank, Pearson Correlation, Kendall Tau) and select to display the best one (Pearson Correlation) on the basis of its average performance on the test set across all experiments. Due to the very poor performance of all metrics when operating on a small number of time-steps ($\leq2$), we only provide the results in Table 1 (avg$\pm$std columns) when these models operate on longer sequences.
    
    \item \texttt{PROCR}, \texttt{PROCR$_*$}, \texttt{GT$_\beta$}, \texttt{RAND}: there are no hyper-parameter to tune in these models.
\end{itemize}

\section{Complete Results on Real Data}\label{app:B}
The complete list of results ($\mu_r$) that were presented in Table~\ref{tab:realresults} are provided in Table~\ref{tab:complete_murank_results}. The interpretation of the ``year'' for each model is provided in Table~\ref{tab:tabtabtabtab}.

\begin{table}[H]
\resizebox{\textwidth}{!}{%
    \centering
    \begin{tabular}{|c|r|r|r|r|r|r|r|r|r|r|r|r|}
    \hline
	year&PROCR&PROCR$_k$&PROCR$_{kt}$&RF&LSTM$_r$&LSTM$_f$&GT$_{\beta}$&	GT$_{c}$&PROCR$_*$&seq2seq$_r$&seq2seq$_f$&seq2seq$_{rf}$\\ \hline
2001&	34.26&	34.11&	34.43&	37.35&	33.67&	36.43&	-&	-&	34.26&	33.66&	23.86&	23.67\\
2002&	32.70&	32.66&	32.41&	34.94&	31.20&	32.98&	-&	-&	32.98&	34.06&	23.52&	23.42\\
2003&	29.24&	29.51&	29.41&	36.94&	30.32&	32.57&	37.59&	43.34&	31.02&	32.44&	23.39&	23.47\\
2004&	25.46&	25.45&	25.03&	27.25&	24.66&	26.08&	35.43&	42.98&	28.68&	30.01&	23.84&	23.50\\
2005&	29.04&	29.10&	28.65&	31.43&	28.98&	29.17&	38.47&	44.47&	28.23&	29.05&	24.21&	23.93\\
2006&	27.73&	28.36&	27.38&	28.86&	26.61&	26.55&	38.74&	44.45&	27.71&	28.58&	24.77&	24.28\\
2007&	26.70&	26.95&	26.64&	30.16&	25.45&	26.39&	34.16&	41.93&	26.98&	28.09&	25.62&	25.17\\
2008&	28.30&	28.23&	27.87&	32.77&	26.25&	27.86&	35.02&	42.86&	26.72&	27.38&	26.53&	24.44\\
2009&	26.10&	26.22&	25.81&	23.27&	24.97&	23.73&	34.23&	43.24&	26.15&	25.71&	27.30&	24.72\\
2010&	27.95&	28.09&	27.38&	28.25&	28.18&	28.19&	36.04&	44.77&	25.81&	25.84&	29.50&	24.83\\
2011&	25.71&	25.91&	25.74&	28.15&	26.07&	26.24&	34.78&	43.99&	25.31&	24.65&	30.91&	25.14\\
2012&	26.77&	27.12&	27.44&	26.51&	27.52&	27.12&	35.18&	44.53&	24.94&	24.42&	33.65&	24.93\\
2013&	30.63&	31.47&	31.91&	30.01&	27.87&	28.62&	38.09&	47.87&	25.01&	24.75&	36.09&	-\\\hline
\textbf{AVERAGE}&	28.51&	28.71&	28.47&	30.45&	27.83&	28.61&	36.16&	44.04&	27.99&	28.36&	27.17&	24.29\\\hline
    \end{tabular}}
    \caption{Complete $\mu_r$ scores across all runs.}
    \label{tab:complete_murank_results}
\end{table}

\begin{table}[H]
\resizebox{\textwidth}{!}{%
\begin{tabular}{lll}
\multicolumn{1}{c}{\textbf{Model}}                                            & \multicolumn{1}{c}{\textbf{Explanation}}                                                              & \multicolumn{1}{c}{\textbf{\begin{tabular}[c]{@{}c@{}}Example (year=2006)\end{tabular}}} \\ \hline \hline

\begin{tabular}[c]{@{}l@{}}PROCR\\ PROCR$_k$\\ PROCR$_{kt}$\end{tabular} &   \begin{tabular}[c]{@{}l@{}}Date to use for aligning the word\\vectors with their corresponding \\ones in the year 2000.\end{tabular} &  \begin{tabular}[c]{@{}l@{}}The model aligns the word vectors in the year\\ 2006 with the word vectors in the year 2000.\end{tabular}\\ \hline

LSTM$_r$ & \begin{tabular}[c]{@{}l@{}}The date indicating the word vectors to\\reconstruct, along with those in the first\\ time-step. \end{tabular} & \begin{tabular}[c]{@{}l@{}}\texttt{LSTM$_r$} receives as input the word vectors in the\\years 2000 and 2006 and reconstructs them.\end{tabular}  \\\hline

\begin{tabular}[c]{@{}l@{}}LSTM$_f$,\\ RF\end{tabular}                   & \begin{tabular}[c]{@{}l@{}}The date indicating the word vectors to\\predict.\end{tabular} &  \begin{tabular}[c]{@{}l@{}}\texttt{LSTM$_f$}/\texttt{RF} receives the word vectors in the year \\2000  \& predicts the word vectors in the year 2006.\end{tabular}\\\hline

\begin{tabular}[c]{@{}l@{}}PROCR$_*$, \\GT$_c$,\\GT$_{\beta}$\end{tabular}&
\begin{tabular}[c]{@{}l@{}}Cut-off date to use for constructing \\the time series of the cosine distances.\end{tabular}
& \begin{tabular}[c]{@{}l@{}}The time series of cosine distances of every word\\are constructed based on the years [2000-2006].\end{tabular} \\ \hline

seq2seq$_r$                                                   & \begin{tabular}[c]{@{}l@{}}Cut-off date in the input, indicating \\ the range of years to reconstruct.\end{tabular} & \begin{tabular}[c]{@{}l@{}}\texttt{seq2seq$_r$} is fed with the word representations \\in  the years [2000-2006] and reconstructs them.\end{tabular} \\\hline

seq2seq$_f$ & \begin{tabular}[c]{@{}l@{}}Cut-off date in the input, affecting \\ the range of years to predict.\end{tabular} &  \begin{tabular}[c]{@{}l@{}}\texttt{seq2seq$_f$} predicts the word vectors during the \\years [2007-2013], given the vectors during the \\ years [2000-2006] as input.\end{tabular} \\\hline

seq2seq$_{rf}$ & \begin{tabular}[c]{@{}l@{}}Cut-off date in the input, indicating \\ the range of years to reconstruct \&\\ affecting the range of dates to predict.\end{tabular} &  \begin{tabular}[c]{@{}l@{}}\texttt{seq2seq$_{rf}$} receives the word vectors during the \\years [2000-2006] and (a) reconstructs them \& (b) \\predicts their representations in [2007-2013].\end{tabular} \\ \hline
\end{tabular}}%
\caption{Explanation of the variable \textit{``year''} in Table~\ref{tab:complete_murank_results}.}
\label{tab:tabtabtabtab}
\end{table}

\end{document}